%% file: main_arxiv.tex
\documentclass[10pt,twocolumn,letterpaper]{article}

\usepackage[pagenumbers]{cvpr}

\input{helpers/packages}
\usepackage[accsupp]{axessibility}
\input{helpers/macros}

\def\confName{CVPR}
\def\confYear{2022}

\begin{document}

\title{Spatially-Adaptive Multilayer Selection for GAN Inversion and Editing}
\author{
Gaurav Parmar $^{1,2}$ \qquad
Yijun Li $^2$ \qquad
Jingwan Lu $^2$ \qquad
Richard Zhang$^2$ \qquad \\
Jun-Yan Zhu$^1$ \qquad 
Krishna Kumar Singh $^2$ \\
$^1$Carnegie Mellon University \qquad $^2$Adobe Research \qquad\\}

\captionsetup[figure]{font=small}
\captionsetup[table]{font=small}

\input{figures/fig1_teaser}

\maketitle

\input{sections/0_abstract}

\input{sections/1_introduction}

\input{sections/2_related_work}
\input{sections/3_method}
\input{sections/4_experiments}
\input{sections/5_discussion}

\bibliographystyle{ieee_fullname}
\bibliography{main}

\clearpage
\input{sections/6_appendix}

\end{document}

%% file: helpers/packages.tex
\usepackage{color}
\usepackage{graphicx}
\usepackage{amsmath}
\usepackage{amssymb}
\usepackage{booktabs}
\usepackage{color}
\usepackage{setspace}
\usepackage{multirow}
\usepackage{array}
\usepackage{tabularx}
\usepackage{longtable}
\usepackage{pdflscape}
\usepackage{pifont}
\usepackage{arydshln}

\usepackage[pagebackref,breaklinks,colorlinks]{hyperref}

%% file: helpers/macros.tex
\newcommand{\reffig}[1]{Figure~\ref{fig:#1}}
\newcommand{\refsec}[1]{Section~\ref{sec:#1}}

\newcommand{\reftbl}[1]{Table~\ref{tab:#1}}

\newcommand{\lblfig}[1]{\label{fig:#1}}
\newcommand{\lblsec}[1]{\label{sec:#1}}

\newcommand{\lbltbl}[1]{\label{tab:#1}}

\newcommand{\myparagraph}[1]{\vspace{0pt}\paragraph{#1}}

\definecolor{mydarkblue}{rgb}{0.0,0.0,.7}
\definecolor{mydarkred}{rgb}{0.8,0.02,0.02}
\definecolor{mydarkorange}{rgb}{0.40,0.2,0.02}

\newcommand{\updates}[1]{{#1}}

\definecolor{mydarkred}{rgb}{0.8,0,0.169}

\newcommand{\IconCheck}{\includegraphics[height=2.8mm, trim=0 2.5mm 0 0mm]{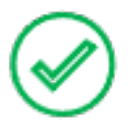}\hspace{0.5mm}}
\newcommand{\IconCross}{\includegraphics[height=2.8mm, trim=0 2.5mm 0 0mm]{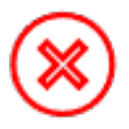}\hspace{0.5mm}}

%% file: figures/fig1_teaser.tex
\twocolumn[{%
\renewcommand\twocolumn[1][]{#1}%
\vspace{-1em}
\maketitle

\vspace{-5mm}
\begin{center}
    \centering 
    \vspace{-5mm}
    \includegraphics[width=\linewidth]{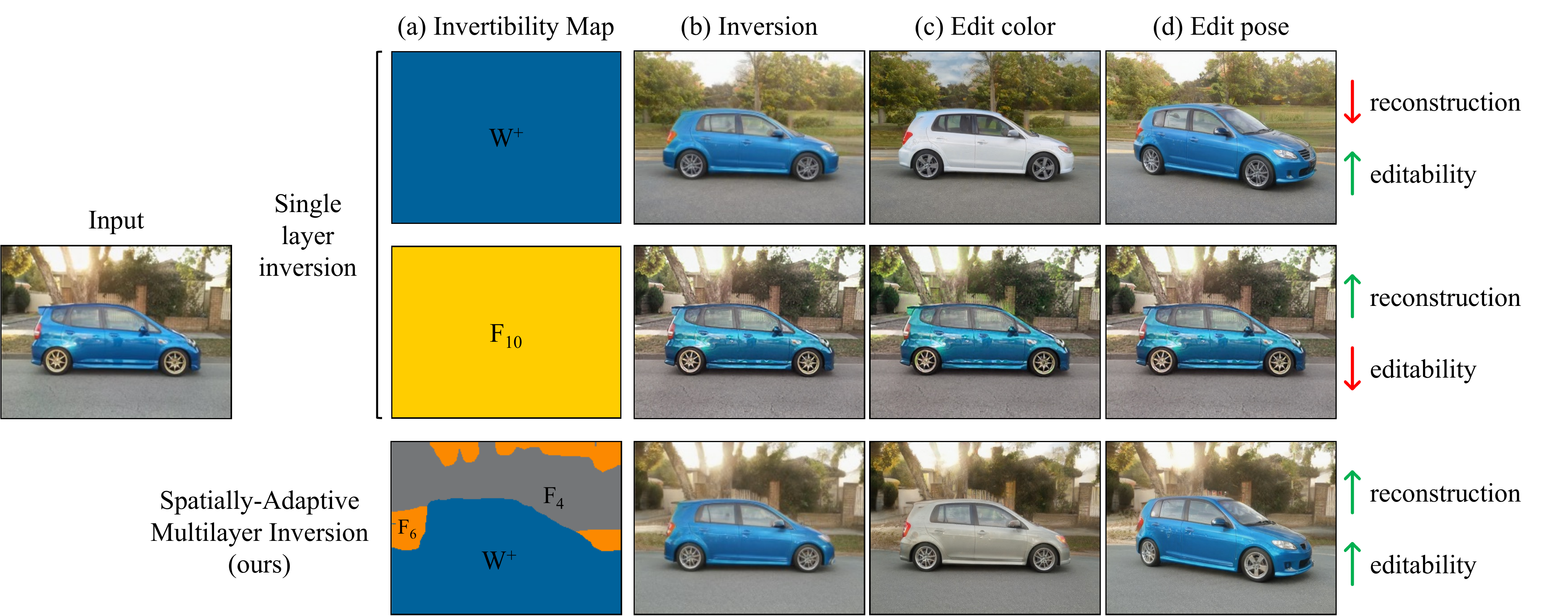}
    \vspace{-5mm}
    \captionof{figure}{\textbf{Inverting and editing an image with spatially adaptive multilayer latent codes.} Choosing a single latent layer for GAN inversion leads to a dilemma between obtaining a faithful reconstruction of the input image and being able to perform downstream edits (1st and 2nd row). In contrast, our proposed method automatically selects the latent space tailored for each region to balance the reconstruction quality and editability (3rd row). Given an input image, our model predicts an invertibility map (a), which contains the layer index used for each region. This allows us to precisely reconstruct the input image (b) while preserving editability (c,d).  }
    \lblfig{teaser}
    \vspace{-2mm}
\end{center}
}]

%% file: sections/0_abstract.tex
\begin{abstract}
\vspace{-3mm}
Existing GAN inversion and editing methods work well for aligned objects with a clean background, such as portraits and animal faces, but often struggle for more difficult categories with complex scene layouts and object occlusions, such as cars, animals, and outdoor images. We propose a new method to invert and edit such complex images in the latent space of GANs, such as StyleGAN2. 
Our key idea is to explore inversion with a collection of layers, \updates{spatially} adapting the inversion process to the difficulty of the image.
We learn to predict the ``invertibility'' of different image segments and project each segment into a latent layer. Easier regions can be inverted into an earlier layer in the generator's latent space, while more challenging regions can be inverted into a later feature space. Experiments show that our method obtains better inversion results compared to the recent approaches on complex categories, while maintaining downstream editability. \updates{Please refer to our project page at \url{https://www.cs.cmu.edu/~SAMInversion}.}
\end{abstract}

%% file: sections/1_introduction.tex
\section{Introduction} 
\lblsec{intro}
\vspace{-2mm}
The recent advances of Generative Adversarial Networks~\cite{goodfellow2014generative}, such as ProGAN~\cite{karras2018progressive}, the StyleGAN model family~\cite{karras2019style,karras2020analyzing,karras2021alias}, and BigGAN~\cite{brock2019large}, have revived the interest in GAN inversion and editing~\cite{zhu2016generative,brock2016neural}. In GAN editing pipelines, one first projects an image into the latent space of a pre-trained GAN, by minimizing the distance between the generated image and an input image. We can then change the latent code according to a user edit, and synthesize the output accordingly. The latent code can then be changed, in order to satisfy a user edit. The final output image is synthesized with the updated latent code. Several recent methods have achieved impressive editing results for real images~\cite{abdal2020image2stylegan++,patashnik2021styleclip,bau2019semantic,zhu2020indomain} using scribbles, text, attribute, and object class conditioning. However, existing methods work well for human portraits and animal faces but are less applicable to more complex classes such as cars, horses, and cats. Compared to faces, these objects have more diverse visual appearance and cluttered backgrounds. In addition, they tend to be less aligned and more often occluded, all of which make inversion more challenging. 

In this work, we aim to invert complex images better. 
We build our method upon two key observations. 

(1) \emph{\updates{Spatially-adaptive} invertibility}: first, the inversion difficulty varies across different regions within an image. Even if the entire image cannot be inverted in the early latent spaces (e.g., $W$ and $W^+$ space of StyleGAN2~\cite{karras2020analyzing}), if we break the image into multiple segments, the easier regions can still be inverted in these latent spaces with high fidelity. For example, in \reffig{teaser}, while the car and sky regions are well-modeled by the \textsc{LSUN Car} generator, shrubs and fences are not, as they appear less frequently in the dataset. Besides, both regions are occluded by the foreground car.

\updates{
(2) \emph{The trade-off between invertibility and editability}: as noted by prior work~\cite{tov2021designing,zhu2020improved}, the choice of layer can determine how precisely an image can be reconstructed and the range of downstream edits that can be performed. Early latent layers of a generative model ($W$, $W^{+}$) are often unable to reconstruct challenging images, but allow meaningful global and local editing. In contrast, inversion using later intermediate layers reconstructs the image more precisely at the cost of reduced editing capability. As invertibility increases in later layers, the editability decreases. The first two rows in \reffig{teaser} show these trade-offs concretely for a real car image.  
}

Considering the spatially-varying difficulty and the trade-off between editability and invertibility, we perform spatially-adaptive multilayer (SAM) inversion by choosing different features or latent spaces to use when inverting each image region. We train a \updates{prediction network} to infer an invertibility map for an input image indicating the latent spaces to be used per segment as shown in the second column of \reffig{teaser}. 
\updates{Our approach enables generating images very close to the target input images while maintaining the downstream editing ability.}

\updates{We conduct experiments on multiple domains such as \textsc{Faces}, \textsc{Cars}, \textsc{Horses}, and \textsc{Cats}. The results show that our method can maintain editability while reconstructing even challenging images more precisely. We measure reconstruction with standard metrics such as PSNR and LPIPS. Whereas, the image quality and the editability are evaluated using a human preference study. Finally, we demonstrate the generality of our idea on different generator architectures (StyleGAN2~\cite{karras2020analyzing}, BigGAN-deep~\cite{brock2019large}), and different paradigms (optimization-based or encoder-based).
Please find our code at \url{https://github.com/adobe-research/sam_inversion}.}

%% file: sections/2_related_work.tex
\input{figures/fig2_training_invertibility}

\section{Related Work} 
\lblsec{related}
\vspace{-1mm}

\myparagraph{GAN inversion and editing.} 
Since the introduction of GANs~\cite{goodfellow2014generative}, several methods have proposed projecting an input image into the latent space of GANs for various editing and synthesis applications~\cite{zhu2016generative,brock2017neural,larsen2016autoencoding,perarnau2016invertible}. This idea of using GANs as a strong image prior was later used in image inpainting, deblurring, compositing, denoising, colorization, semantic image editing, and data augmentation~\cite{yeh2017semantic,asim2018blind,gu2020image,bau2019semantic,wu2021towards,chai2021using,chai2021ensembling}. See a recent survey~\cite{xia2021gan} for more details. 
The enormous progress of large-scale GANs~\cite{karras2018progressive,karras2019style,karras2020analyzing,karras2020training,brock2019large,zhao2020diffaugment,karras2021alias,karnewar2020msg}
allows us to adopt GAN inversion for high-resolution images~\cite{abdal2019image2stylegan,abdal2020image2stylegan++}. One popular application is portrait editing~\cite{alaluf2021matter,abdal2021styleflow,tewari2020stylerig,luo2020time}.

Current methods can be categorized into three groups: optimization-based, encoder-based, and hybrid methods.
The optimization-based methods~\cite{zhu2016generative,lipton2017precise,karras2020analyzing,abdal2019image2stylegan,abdal2020image2stylegan++} aim to minimize the difference between the optimization output and the input image. 
Despite achieving fairly accurate results, the slow process requires many iterations and may get stuck in local optimum. To accelerate the process, several works~\cite{larsen2016autoencoding,zhu2016generative,brock2017neural,perarnau2016invertible,tov2021designing,richardson2020encoding,wei2021simpleinversion} learn an encoder to predict the latent code via a single feed-forward pass. %
However, the learned encoder is sometimes limited in reconstruction quality compared to the optimization-based scheme.
Naturally, hybrid approaches that combine the best of both schemes emerge~\cite{zhu2016generative,alaluf2021restyle,bau2019seeing,bau2019semantic,huh2020transforming, wei2021simpleinversion}, but the trade-off between quality and speed still persists. 

\vspace{-5mm}
\myparagraph{Choosing the latent space.} Several previous methods~\cite{abdal2019image2stylegan,abdal2020image2stylegan++} focus on inverting the input image into the latent space of StyleGAN models~\cite{karras2019style,karras2020analyzing} that use AdaIN layers~\cite{huang2017adain} to control the ``style'' of an image. In addition to exploring different projection schemes, they demonstrate that the choice of latent space is a key factor due to the unique style-based design of the StyleGAN. Instead of projecting an image into the latent space~\cite{zhu2016generative,brock2017neural}, recent works propose  projecting an image into style parameter space~\cite{abdal2019image2stylegan,abdal2020image2stylegan++,wu2020stylespace} and convolutional feature space~\cite{zhu2021barbershop}. As noted by recent work~\cite{tov2021designing,zhu2020improved}, there exists a trade-off between the invertibility and editability, and no layer can maximize both criteria at the same time. 

    To handle complex images, recent papers propose using multiple codes of the same layer~\cite{gu2020image,kafri2021stylefusion,suzuki2018spatially}, \updates{splitting image into segments~\cite{futschik2021real}, using consecutive images~\cite{xu2021continuity}}, explicitly handling misaligned objects~\cite{huh2020transforming,anirudh2020mimicgan,kang2021gan}, modifying the generator architecture for better editing ability~\cite{kim2021exploiting,park2020swapping}, adopting a class-conditional GAN~\cite{huh2020transforming,pan2021exploiting,suzuki2018spatially}, and %
fine-tuning the generator to an input  image~\cite{bau2019semantic,roich2021pivotal,pan2021exploiting}. 

Different from the above methods that operate on a single layer, 
we take into account the inversion difficulty across different input image segments and perform the inversion separately for each segment by using multiple latent spaces. 
We show that our method outperforms a concurrent generator fine-tuning method~\cite{roich2021pivotal} in our experiments. 

\vspace{-5mm}
\myparagraph{Finding editing directions.} After inversion, we can edit the inverted code by traversing semantically meaningful directions computed using supervised~\cite{gansteerability,bau2019gandissect,shen2020interfacegan} or unsupervised approaches~\cite{shen2021closed,harkonen2020ganspace,peebles2020hessian,voynov2020unsupervised,collins2020editing}. Most of these methods compute these directions offline~\cite{gansteerability,bau2019gandissect,shen2021closed} and provide them as pre-canned options for users. Other works calculate the editing directions during inference time to support more flexible editing interfaces with scribbles~\cite{zhu2016generative} and text inputs~\cite{patashnik2021styleclip}. We show that our method can work well with different types of directions. 

%% file: figures/fig2_training_invertibility.tex
\begin{figure*}[t]
    
    \centering
    \includegraphics[width=0.99\linewidth]{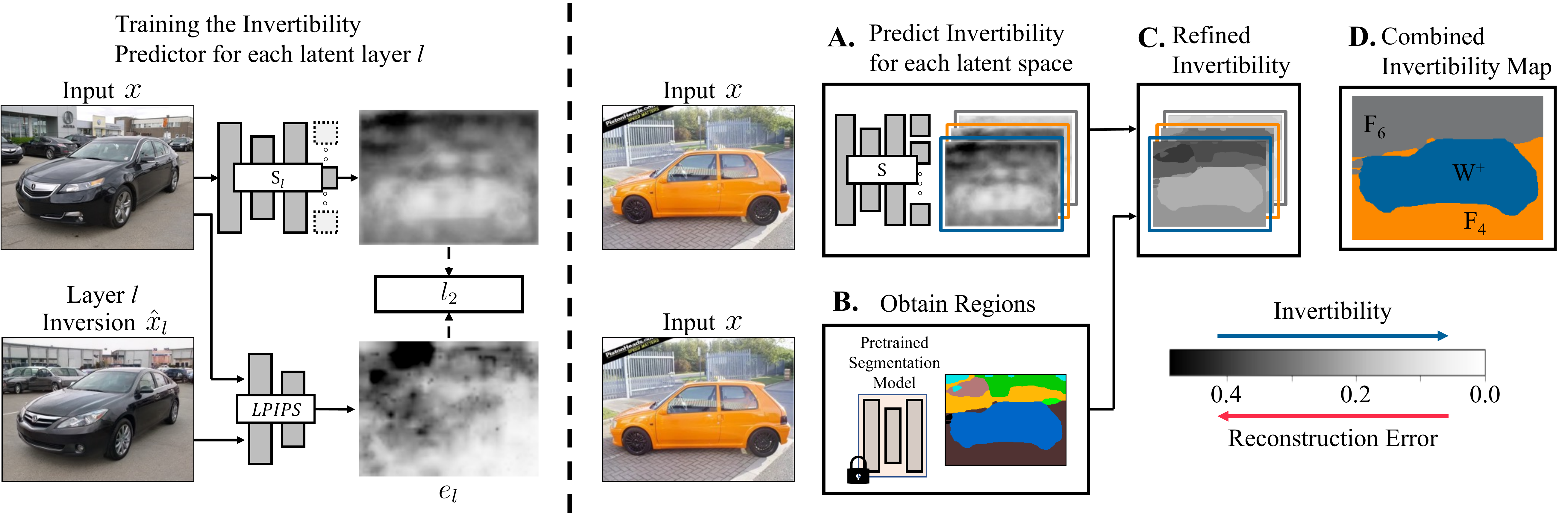}
    \vspace{-2.5mm}
    \caption{
    \textbf{Training the Invertibility Segmenter.} On the left we show how each of the invertibility predictior $S_l$ are trained. We invert all images in the training set using one of the five candidate latent spaces and use the LPIPS ~\cite{zhang2018unreasonable} spatial error map $e_l$ as supervision. 
    Next (right) we show how the trained invertibility models are used to generate the final inversion latent map. We first predict how difficult each region of the image is to invert for every latent layer using our aforementioned invertibility network. Subsequently we refine the predicted map using a pre-trained semantic segmentation network and combine them using the user-specified threshold $\tau$. This combined invertibility map shown on the right is used to determine the latent layer to be used for inverting each segment in the image. }
    \lblfig{inv_train}
    \vspace*{-5mm}
\end{figure*}

%% file: sections/3_method.tex
\vspace{-1mm}
\section{Approach} 
\lblsec{approach}
\vspace{-1mm}
We aim to invert images using a pretrained GAN while maintaining editability. We begin by learning to predict an invertibility map that indicates which latent spaces should be used for each image region. Next, we fuse features from different latent spaces to generate an image that matches our input and can be edited in the latent space. 

\vspace{-1mm}
\subsection{Predicting Invertibility}
\vspace{-2mm}
As discussed previously, different latent spaces have different inversion capabilities. We learn a network to predict what parts of the image are invertible using any given latent space. Here we use ``invertibility'' to indicate how closely our generated result can match the input image. In \reffig{inv_train} (left), we show how we learn invertibility predictor for different latent spaces. We collect a dataset of image pairs that consists of the input image $x \in R^{H\times W\times 3}$ and its reconstruction $\hat{x}_l \in R^{H\times W\times 3}$ into the $l^\text{th}$ latent space, following the optimization-based inversion suggested by Karras et al.~\cite{karras2020analyzing}.
We consider 5 different latent spaces $\Phi = \{W^{+}, F_{4}, F_{6}, F_{8}, F_{10} \}$, where the index of $F$ corresponds to the feature layer index of the StyleGAN2 generator and $W^{+}$ is the concatenation of different vectors from $W$ space, in which $W$ space is the output space of the MLP network of StyleGAN2. 
We choose $W^{+}$ instead of $W$, as it provides better inversion results and more fine-grained and disentangled control when performing the downstream edits. Next, we compute the reconstruction loss as follows
\vspace{-1mm}
\begin{equation}
    e_l =\mathcal{L}_{\text{LPIPS}}(x, \hat{x}_l),
\end{equation}
where $e_l  \in R^{H\times W}$ is the LPIPS spatial error map~\cite{zhang2018unreasonable} between $x$ and their inversions $\hat{x}_l$ for each latent space. 

The parts that are easy to invert have smaller spatial errors, whereas difficult regions induce larger errors. We subsequently train a network to predict the invertibility for each latent space,  regressing to the LPIPS spatial error map via an $\ell_2$ loss. The training loss can be formulated as follows: 
\vspace{-2mm}
\begin{equation}
S_l=\arg\min_{S_l} \ell_2 \big(S_l(x), e_l\big).
\vspace{-2mm}
\end{equation}
Once trained, this network predicts the invertibility for any input image, at any layer, in a feed-forward fashion. 
However, our prediction can be noisy and may not be consistent within the same semantic region. This could potentially result in inconsistent inversions and edits, as different parts of the same region can be assigned to different latent codes. We refine our prediction using a pretrained segmentation model. For every segment,  we compute the average predicted invertibility in the region and use the value for the entire segment. As shown in \reffig{inv_train} (right), such a refining step helps us align the invertibility map with natural object boundaries in the image.

\input{figures/fig4_tradeoffs}

\vspace{-1mm}
\subsection{Adaptive Latent Space Selection}
\vspace{-2mm}
We observe that latent spaces have an inherent trade-off between reconstructing the input image and utility for downstream image editing tasks, as also noted by recent work~\cite{tov2021designing,zhu2020improved}. For example, choosing the latent space to be $W^{+}$ would result in an inverted latent vector that is amenable for editing, but sub-optimal for obtaining a faithful reconstruction for difficult input images. On the other hand, choosing activation block $F_{10}$ (close to the generated pixel space) would have great reconstruction, but limited editing ability. In \reffig{latent_tradeoffs}, we show this trade-off for different choices of latent spaces explicitly. We invert the input image using a single latent layer, and observe that the reconstruction quality improves monotonically as we use layers increasingly closer to the output pixels.

Committing to a \textit{single} latent layer for the whole image forces us to a single operating point on the trade-off between editability and reconstruction, across the whole image. Instead, we aim to \textit{adapt} the latent layer selection, depending on the image content in a region.
To do this, for each image segment, we choose the earliest latent layer, such that the reconstruction still meets some minimum criteria.

More concretely, for each segment, we choose the most editable latent space from $\Phi$ ($W^{+}$ being most editable and $F_{10}$ being least), with predicted invertibility above threshold $\tau$ for that segment. We choose this threshold value empirically such that the inversion is perceptually close to the input image, without severely sacrificing editability. In \reffig{method}, we show our final inversion map, with different latent spaces assigned to different segments in the input image. The simple car region gets assigned to the $W^{+}$ space, whereas the difficult to generate background regions, which typically cannot be generated by the native latent space, gets assigned to the later $F_4$ and $F_6$ latent spaces.

\input{figures/fig3_image_formation}

\input{figures/fig5_results}
\input{figures/fig6_biggan}

\vspace{-1mm}
\subsection{Training Objective}
\vspace{-2mm}
\updates{We implement our multilayer inversion in two settings: 1) optimization-based and 2) encoder-based. In the optimization-based approach, we directly optimize the latent space $\phi$ for each image. For the encoder-based approach, we train a separate encoder for each latent space. 
}

\vspace{-4mm}
\myparagraph{Image formation model.} 
In \reffig{method}, we show how the latent codes are combined to generate the final image. Our predicted $w^+ \in W^{+}$ is directly used to modulate the layers of pre-trained StyleGAN2. For feature spaces $F \in \{F_4, F_6, F_8, F_{10}\}$, we predict the change in values $\Delta  f$ for the regions that are to be inverted in that layer. We predict the change in layer's feature, rather than directly predicting the feature itself, as propagating the features from earlier layers provides a meaningful initialization to adjust from.

The output feature value is a combination of both $w^+$ and $\Delta f$ masked by a binary mask indicating which region should be inverted in that layer. For example, to produce the feature $f_4 \in F_4$, we have: 
\vspace{-1mm}
\begin{equation}
    f_4=g_{0\rightarrow4}(c,w^+) +m_4 \odot \Delta f_4, 
\end{equation}
where $g_{i\rightarrow j}$ denotes the module from the $i$-th to the $j$-th layers in the convolutional layers of the StyleGAN2, $c$ is the input constant tensor used in StyleGAN2, $m_4$ is the refined, predicted invertibility mask bilinearly downsampled to corresponding tensor size, and $\odot$ denotes the Hadamard product. Note that $g_{i\rightarrow j}$ is modulated by the corresponding part of the extended latent code $w^+$. Similarly, we can calculate all the features and the final output image as follows: 
\begin{align}
f_6&=g_{4\rightarrow 6}(f_4,w^+) +m_6 \odot \Delta f_6 \nonumber \\ 
f_8&=g_{6\rightarrow8}(f_6,w^+) + m_8 \odot \Delta f_8 \nonumber \\
f_{10}&=g_{8\rightarrow10}(f_8,w^+) + m_{10} \odot \Delta f_{10} \nonumber \\
\hat{x}&=g_{10\rightarrow 16}(f_{10},w^+).
\end{align}
Next, we present our objective functions to optimize the latent code $\phi = \{w^+, \Delta f_4, \Delta f_6, \Delta f_8, \Delta f_{10} \}$. We reconstruct the input image while regularizing the latent codes.

\vspace{-4mm}
\myparagraph{Reconstruction losses.} 
We use the $\mathcal{L}_2$ distance between the inverted image $\hat{x}$ and the input image $x$ along with LPIPS difference as our reconstruction losses. 
\vspace{-1mm}
\begin{equation}
    \mathcal{L}_{\text{rec}} = \ell_2 (x, \hat{x}) + \lambda_{\text{LPIPS}} \mathcal{L}_{\text{LPIPS}}(x, \hat{x}),
\end{equation}
where $\lambda_{\text{LPIPS}}$ is the weight term.

\vspace{-4mm}
\myparagraph{$W$-space regularization.} 
As noted in \cite{tov2021designing, wulff2020improving}, inverting an image with just reconstruction losses results in latent codes that are not useful for editing. For our inversion methods, we use different latent regularization losses for different latent spaces. For $w^{+}$, we use the following: 
 \begin{equation}
\mathcal{L}_{W}=\sum_n^N \big [ (\hat{w}_n-\mu)^T \Sigma (\hat{w}_n-\mu) + ||w_n^+ -w_0^+||^2 \big ], 
\end{equation}
where $w_n^+$ is the $n^\text{th}$ component of the $w^+$ vector, $\hat{w}_n  = \texttt{LeakyReLU} (w_n^+, 5.0)$,  $\mu$ and $\Sigma$ are the empirical mean and covariance matrix of randomly sampled and converted $\hat{w}$ vectors respectively. The first term applies a Mutlivariate Gaussian prior~\cite{wulff2020improving}, and the second term minimizes the variation between the individual style codes and the first style code. %
\vspace{-4mm}
\myparagraph{$F$-space regularization.} 
For the feature space, we enforce our predicted change $\Delta f$ to be small, so that our final feature value does not deviate much from the original value.
\begin{equation}
    \mathcal{L}_F= \sum_{{\Delta f}\in \phi \backslash w^+ }||\Delta f||^2
\end{equation}

\vspace{-3.5mm}
\myparagraph{Final objective.} Our full objective is written as follows: 
\begin{equation}
\arg \min_{\phi}  \mathcal{L}_{\text{rec}} + \lambda_{W} \mathcal{L}_{W} + \lambda_F \mathcal{L}_{F},
\end{equation}

\vspace{-2mm}
\noindent where  $\lambda_{W}$ and $\lambda_F$ control the weights for each term. 

\subsection{Image Editing}
\vspace{-2mm}
After obtaining the inverted latent codes, we edit the images by applying the edit direction vector to the inverted $w^{+}$ latent vector. We use GANSpace~\cite{harkonen2020ganspace} and StyleCLIP~\cite{patashnik2021styleclip} for finding an editing direction $\delta w^+$ in the $W^{+}$ latent space. Segments inverted in $W^{+}$ space get modulated by the entire code $w^{+}+\delta w^+$, whereas segments inverted in intermediate feature spaces $\{ F_4, F_6, F_8, F_{10} \}$ get modulated by $w^{+}+\delta w^+$ only for the layers which come after that feature space layer. For example, segments inverted in $F_{10}$ space get modulated by $w^{+}$ for the layers until the $10^\text{th}$ layer, and  $w^{+}+\delta w^{+}$ for the layers afterward. This is necessary, as our inverted feature would not be compatible with $w^{+}+\delta  w^+$. 

\input{figures/fig7_runtime}
\input{figures/fig7_cmp_baselines}

%% file: figures/fig4_tradeoffs.tex
\begin{figure}[t!]
    \centering
    \includegraphics[width=\linewidth]{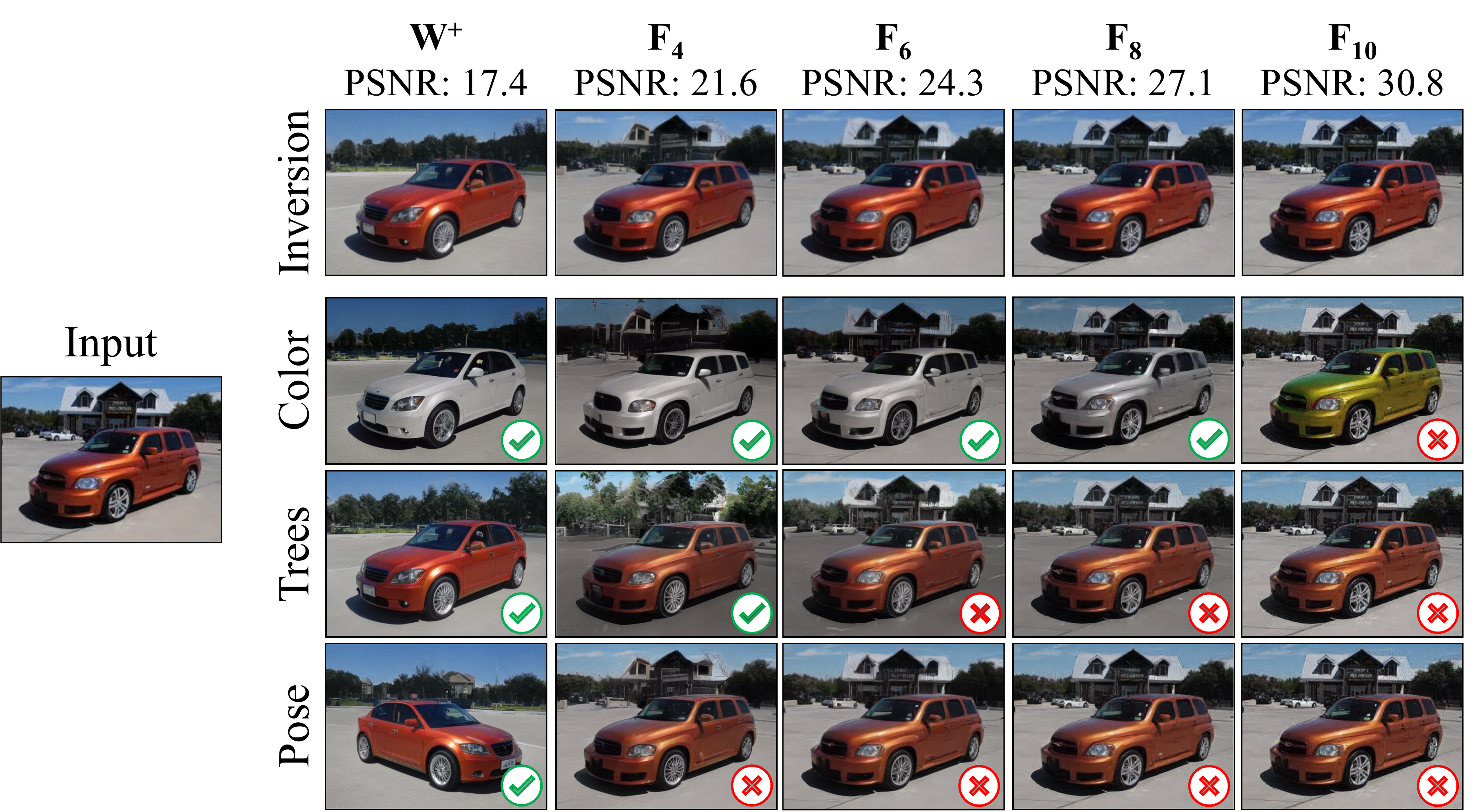}
    \vspace{-5mm}
    \caption{
    \updates{
    \textbf{Trade-offs between invertibility and editibility.} 
    We show inversion and editing when the input is inverted using different \textit{single} latent layers. As we go down in feature space reconstruction improves but editing capabilities decreases. The improvement in reconstruction is shown visually for a single image and quantitatively with PSNR using 1000 images. Whether the edit was applied successfully is indicated by \IconCheck and \IconCross. }
    }
    \lblfig{latent_tradeoffs}
   \vspace*{-5mm}
\end{figure}

%% file: figures/fig3_image_formation.tex
\begin{figure}[t!]
    \centering
    \includegraphics[width=0.99\linewidth]{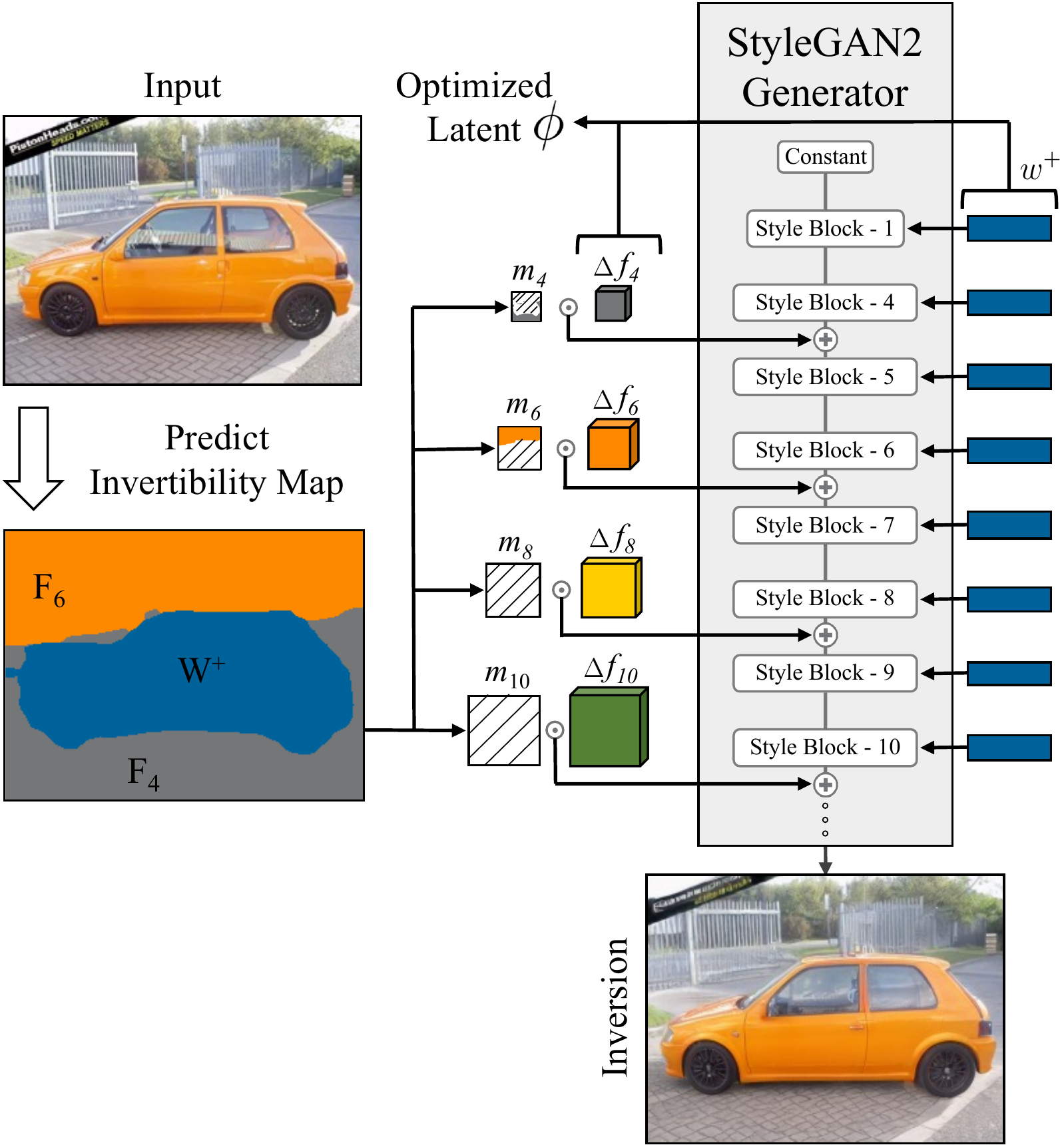}
    \vspace{-2mm}
    \caption{
    \updates{\textbf{Image formation using spatially adaptive latent codes.} 
    We show how the predicted invertibility map is used in conjunction with multiple latent codes to generate the final image. 
    $w^+ \in W^+$ directly modulates the StyleBocks of the pretrained StyleGAN2 model. For intermediate feature space $F_i$, we predict the change in layer's feature value $\Delta f_i$ and add it to the feature block after masking with the corresponding  binary mask $m_i$. 
    }
    }
    \lblfig{method}
    \vspace{-5mm}
\end{figure}

%% file: figures/fig5_results.tex
\begin{figure*}[t]
    \centering
    \includegraphics[width=1.0\linewidth]{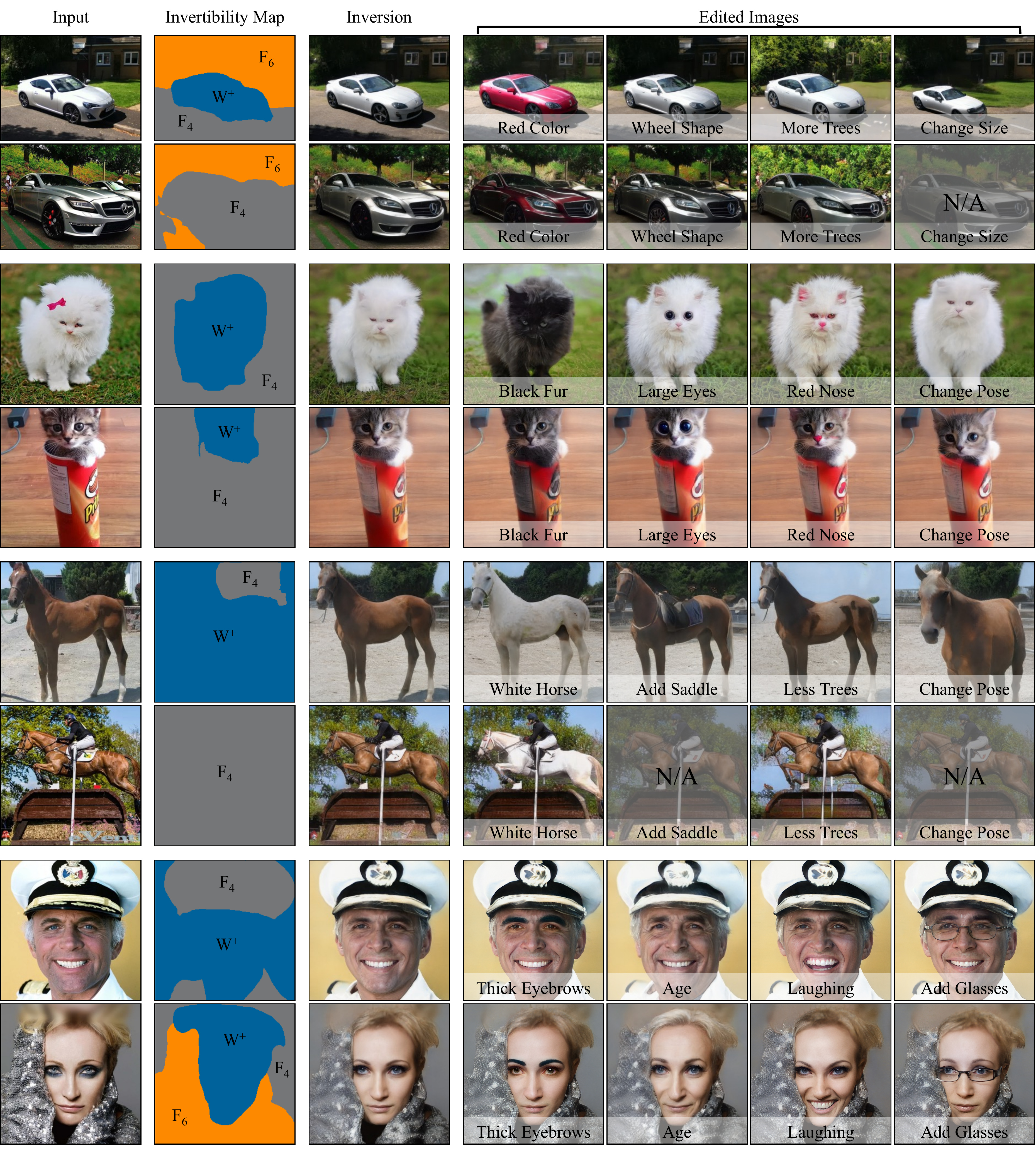}
    \vspace{-8mm}
    \caption{
    \textbf{Qualitative inversion and editing results.} In the first column we show input images for which we predict the invertibility map shown in the second column. 
    We are able to obtain inverted images which closely match the input as shown in third column. In the remaining columns, we show our edit results. We can apply complex spatial edits like pose and size changes in seamless fashion even though different segments were inverted in different latent spaces.   
    }
    \lblfig{qual_results}
    \vspace*{-5mm}
\end{figure*}

%% file: figures/fig6_biggan.tex
\begin{figure*}[t]
    \centering
    \includegraphics[width=\linewidth]{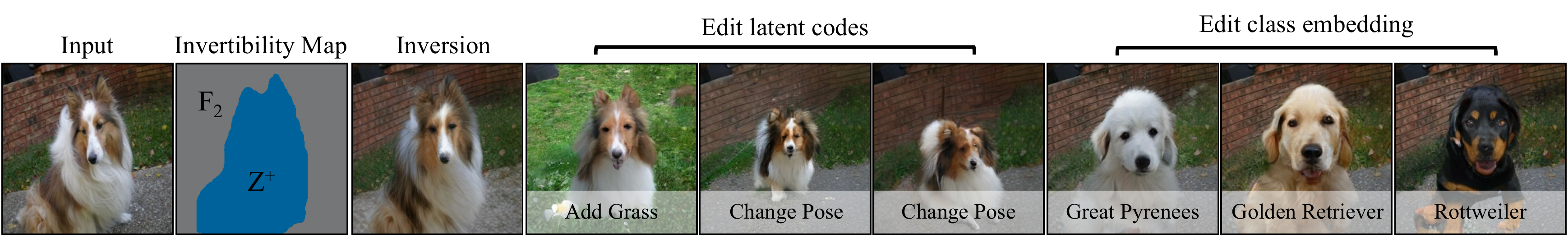}
    \vspace{-7mm}
    \caption{
    \updates{
    \textbf{Inversion and editing using BigGAN-deep.} We show that our spatially-adaptive method of using different latent layers ($\text{Z}^+, \text{F}_2$) can be applied to class-conditional models such as BigGAN-deep ~\cite{brock2019large} trained on ImageNet. In the third column we show that the inversion obtained is very close to the input image. Subsequent edits can be performed using either changing the latent code (top row) or modifying class embedding vector (bottom row). 
    }}
    \lblfig{biggan_results}
    \vspace*{-5mm}
\end{figure*}

%% file: figures/fig7_runtime.tex
\begin{figure}[t]
    \centering
    \includegraphics[width=1.00\linewidth]{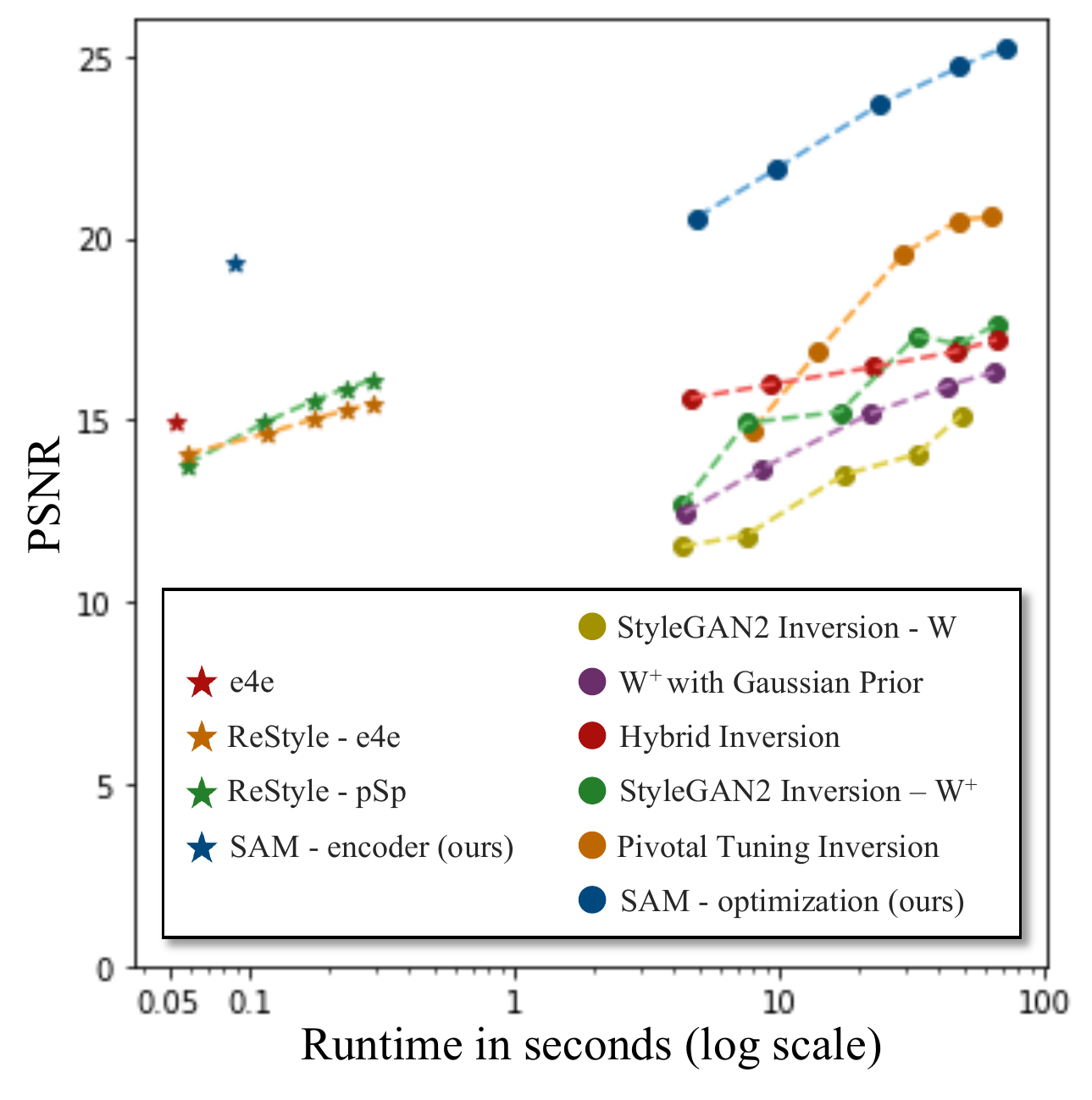}
    \vspace{-9mm}
    \caption{
    \textbf{Reconstruction at different runtimes. } 
    We compare the reconstruction of different GAN inversion methods in the optimization and encoder regimes using 1000 car images. Each of the method uses a single NVIDIA RTX 3090 GPU. Our proposed method achieves a closer reconstruction to the input in a shorter amount of time for both the optimization and encoder paradigms.}
    \lblfig{runtime_comparison}
    \vspace{-5mm}
\end{figure}

%% file: figures/fig7_cmp_baselines.tex
\begin{figure*}[t]
    \centering
    \includegraphics[width=1.0\linewidth]{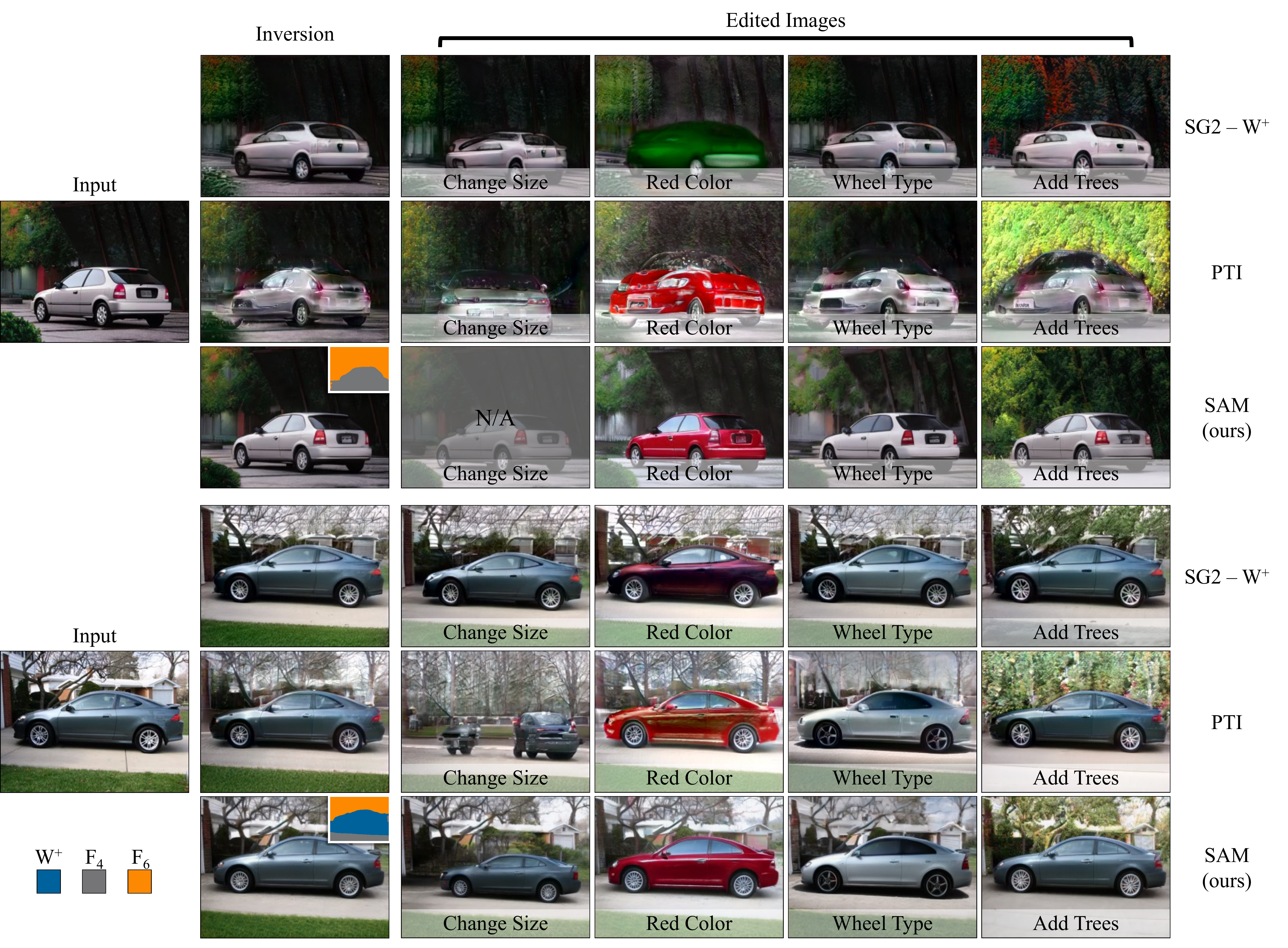}
    \vspace{-6mm}
    \caption{
    \textbf{Comparison with other optimization based inversion methods.} We compare our inversion and editing results with StyleGAN2 $W^{+}$ inversion and pivotal tuning. We obtain much closer and detailed inversion to the target image compared to other approaches.  Also, we are able to apply semantic edits while maintaining the realism of the image. We are able to perform both low level edits like color change as well as high level edits like size changes.  Additional results on other categories are shown on the project website. 
    }
    \lblfig{cmp_related}
    \vspace*{-.1in}
\end{figure*}

%% file: sections/4_experiments.tex
\section{Experiments} \lblsec{experiments}
\lblsec{exp} 
Here we perform detailed quantitative and qualitative analysis to show effectiveness of our inversion method across different datasets. \updates{Please refer to the appendix for additional details including BigGAN inversion details, LPIPS architecture variations, more qualitative results, face editing experiments, and ablation studies. }

\vspace{-3.5mm}
\myparagraph{Datasets.} We test our method on pretrained StyleGAN2 and BigGAN-deep generators trained on a variety of different challenging domains and follow the commonly used protocol for the different domains ~\cite{richardson2020encoding,alaluf2021restyle,roich2021pivotal}. 
For all experiments we use the official released StyleGAN2~\cite{karras2020analyzing} trained on LSUN Cars, LSUN Horses, LSUN Cats, and FFHQ ~\cite{karras2019style} datasets, and the official released BigGAN-deep~\cite{brock2019large} trained on ImageNet~\cite{russakovsky2015imagenet}. We use a subset of 10,000 images from the dataset for training our invertibility prediction network $S$ and 1000 images for the evaluation.

\vspace{-4mm}
\myparagraph{Evaluation.}
We evaluate the performance of various inversion methods on two tasks - reconstruction and editibility. The reconstruction between the inverted image and the input image is measured using PSNR and LPIPS~\cite{zhang2018unreasonable}. Note that different prior inversion methods use different LPIPS backbones. We use LPIPS-VGG for all of our experiments and comparisons. 
As pointed out by \cite{tov2021designing}, measuring the editing ability of the latent codes is difficult and image quality metrics such as IS~\cite{salimans2016improved}, FID~\cite{heusel2017gans} and KID~\cite{binkowski2018demystifying} do not correlate with the user preference. Therefore we show qualitative comparisons and conduct user preference studies to evaluate the quality of inverted and edited images. 
\input{tables/tab1_reconstruction}

\vspace{-4mm}
\myparagraph{Reconstruction comparison.} We first compare our inversion method to other state-of-the-art GAN inversions methods in the optimization-based regime. \emph{StyleGAN2 Inversion} and \emph{StyleGAN2 Inversion using $W^{+}$} invert image in  $W$ and $W^{+}$ latent space respectively. \cite{wulff2020improving} applies multi-variant Gaussian prior constraint while doing the inversion. We also compare against \emph{Hybrid $W^+$ Inversion} that uses a pretrained \emph{e4e} encoder \cite{tov2021designing} for initialization. Recently proposed pivotal tuning inversion (\emph{PTI})~\cite{roich2021pivotal} additionally finetunes the weights of pre-trained StyleGAN2 after inverting the image in the $W$ space. \reftbl{inv_comparison} shows that our method achieves better reconstruction across all the metrics compared to baselines. Our approach is able to invert difficult regions using intermediate layers feature space, whereas baselines struggle to invert by just relying on single $W$ and $W^{+}$ space. \emph{PTI} has the ability to change the StyleGAN2 weights to invert the image, but it uses heavy locality regularization to discourage the deviation from original weights, which limits its inversion capability. Also, for simpler image parts, we get better inversion as our $W^{+}$ latent code just focuses on parts that it can invert. In contrast,  other approaches try to invert both easy and difficult parts using the same code, resulting in sub-optimal inversion even for the easier part. \updates{We perform similar comparisons of encoder based methods and show that an encoder trained using our proposed method outperforms the encoder baselines ~\cite{tov2021designing, alaluf2021restyle} on challenging images. 
On faces, our encoder obtains a similar reconstruction with just a single forward pass as the best performing baseline \emph{ReStyle (pSp)}, which requires five forward passes.}
\updates{We also compare the runtime of optimization-based and encoder-based inversion methods using 1000 Car images in \reffig{runtime_comparison}. In both paradigms, our method obtains a better reconstruction in a shorter amount of time. } 

\vspace{-4mm}
\myparagraph{Qualitative results.} Next, we show our ability to edit the reconstructed complex images in \reffig{qual_results}.  
\updates{In the third column, we show our ability to reconstruct difficult regions using more capable latent layers $F_4 ~\text{and}~ F_6$, whereas the easy-to-generate regions use the more editable $W^{+}$. }
This separation allows us to perform challenging edits while faithfully reconstructing the target image. 
\updates{\reffig{biggan_results} shows inversion and editing results for a class-conditioned BigGAN model. 
}
In \reffig{cmp_related}, we observe that we get much closer inversion and realistic edits than baselines approaches. In some cases such as the first image, we can preserve even fine-grained details like the type of light and car wheels during editing stage. \emph{StyleGAN2 inversion using $W^{+}$} generates realistic looking images but does not matches the input images well whereas \emph{PTI} generates images that are closer but lack realism,  especially after editing. We hypothesize that this is due to the incompatibility between the finetuned weights and the edit directions learned before finetuning. 

\vspace{-5mm}
\myparagraph{User study.} 
\updates{We additionally conduct a user preference study to evaluate the realism of inverted and edited images. \reftbl{user_study} compares our method to three closest baselines methods (\emph{PTI}~\cite{roich2021pivotal}, \emph{StyleGAN2 Inversion using $W^{+}$}, and \emph{Hybrid $W^+$ Inversion}) using 500 different target images from each category. Every pair is evaluated by 3 randomized and different users, resulting in 1500 comparisons per baseline per category. The results show that users prefer our results over the baselines for all challenging image categories.
}

\input{tables/tab2_mturk}

%% file: tables/tab1_reconstruction.tex
\begin{table*}[t!]
    \centering
    \resizebox{\linewidth}{!}{
    \begin{tabular}{l  cc  cc  cc  cc}
        \toprule 
        \multirow{2}{*}{\textbf{Method}} 
        &\multicolumn{2}{c}{\textbf{Cars}} & \multicolumn{2}{c}{\textbf{Horses}} & \multicolumn{2}{c}{\textbf{Cats}} & \multicolumn{2}{c}{\textbf{Faces}}  \\
        \cmidrule(lr){2-3} \cmidrule(lr){4-5}
        \cmidrule(lr){6-7} \cmidrule(lr){8-9}
        & LPIPS ($\downarrow$)  & PSNR ($\uparrow$) 
        & LPIPS ($\downarrow$)  & PSNR ($\uparrow$) 
        & LPIPS ($\downarrow$)  & PSNR ($\uparrow$)
        & LPIPS ($\downarrow$)  & PSNR ($\uparrow$)\\
        \midrule

        StyleGAN2 Inversion \cite{karras2020analyzing} &
        0.34 & 14.44 & 0.45 & 13.46 & 0.44 & 14.47 & 0.28 & 18.32 \\
        
        StyleGAN2 Inversion \cite{karras2020analyzing} using $W^+$  &
        0.24 & 17.29 & 0.34 & 15.74 & 0.35 & 17.11 & 0.20 & 22.10 \\
        
        Inversion with a Gaussian Prior \cite{wulff2020improving} & 
        0.45 & 15.92 & 0.42 & 17.19 & 0.49 & 17.01 & 0.15 & 25.18 \\
        
        Hybrid $W^+$ Inversion with e4e \cite{tov2021designing} 
            & 0.36 & 17.05 
            & 0.42 & 16.68
            & 0.42 & 17.91
            & 0.15 & 25.13\\
        
        PTI \cite{roich2021pivotal} &
        0.38 & 19.39 & 0.43 & 18.73 & 0.41 & 20.45 & 0.26 & 22.36 \\
        
        SAM - optimization (ours) & 
        \textbf{0.16} & \textbf{22.81} & \textbf{0.23} & \textbf{21.07} & \textbf{0.22} & \textbf{22.91} & \textbf{0.13} & \textbf{26.89} \\
        
        \cdashline{1-9}[4pt/2pt]
        e4e ~\cite{tov2021designing} & 
        0.47 & 14.57 & 0.55 & 13.98 & 0.56 & 14.68 & 0.34 & 19.39 \\
        
        ReStyle (pSp) ~\cite{alaluf2021restyle} &
        0.43 & 16.44 & 0.45 & 16.53 & 0.48 & 17.58 & \textbf{0.29} & \textbf{21.47} \\
        
        ReStyle (e4e) ~\cite{alaluf2021restyle} &
        0.45 & 15.61 & 0.52 & 14.50 & 0.53 & 15.64 & 0.34 & 19.72 \\
        
        SAM - encoder (ours)
            & \textbf{0.28} & \textbf{19.21} 
            & \textbf{0.34} & \textbf{18.61}
            & \textbf{0.37} & \textbf{18.59} & \textbf{0.29} & 21.10 \\
        
        \bottomrule 
        
    \end{tabular}}
    \vspace{-1mm}
    \caption{\textbf{Reconstruction comparison to prior methods.} We use PSNR and the LPIPS-VGG for the evaluating the reconstruction using 1000 images. 
    For the challenging categories, we achieve a better reconstruction than all baseline approaches in both the optimization based and encoder based paradigms. The faces images are simpler and contain fewer challenging regions. Subsequently, our method performs slightly better than prior methods when inverting with optimization and similar to the best performing ReStyle (pSp) with encoders. 
    }
    \lbltbl{inv_comparison}
    \vspace{-5mm}
\end{table*}

%% file: tables/tab2_mturk.tex
\begin{table}[t!]
    \centering
    \resizebox{1.0\linewidth}{!}{
    \begin{tabular}{l ccc ccc }
        \toprule 
        \multirow{2}{*}{\textbf{Method}}& \multicolumn{3}{c}{\textbf{Inversion}} & \multicolumn{3}{c}{\textbf{Editing}} \\
        \cmidrule(lr){2-4}
        \cmidrule(lr){5-7}
        & 
        \textbf{Cars} & \textbf{Horses} & \textbf{Cats} &
        \textbf{Cars} & \textbf{Horses} & \textbf{Cats}\\
        \midrule
        
        PTI \cite{roich2021pivotal} & 7.0\% & 11.6\% & 11.6\% & 18.4\% &16.4\% & 38.0\% \\
        SAM (ours) & 
        \textbf{93.0\%} & \textbf{88.4\%} & \textbf{88.4\%} &
        \textbf{81.6\%} & \textbf{83.6\%} & \textbf{62.0\%} \\
        
        \cdashline{1-7}[4pt/2pt]
        SG2-W+ & 
        28.0\%&24.7\%&20.5\%&
        28.8\%&35.7\%&35.0\%\\
        SAM (ours) & 
        \textbf{72.0\%} & \textbf{75.3\%} & \textbf{79.5\%} & \textbf{71.2\%} & \textbf{64.3\%} & \textbf{65.0\%}\\
        
        \cdashline{1-7}[4pt/2pt]
        e4e hybrid & 
        23.2\% & 21.8\% & 22.4\% & 36.4\% & 38.3\% & 44.6\%\\ 
        SAM (ours) & 
        \textbf{76.8\%} & \textbf{78.2\%} & \textbf{77.6\%} & 
        \textbf{62.6\%} & \textbf{61.7\%} & \textbf{55.4\%}\\
        \bottomrule 
    \end{tabular}
    }
    \vspace{-2mm}
    \caption{\updates{
    \textbf{User preference comparison  with prior methods.} We invert and edit 500 from each of the image categories and ask 3 different users (1500 pairs per comparison). Results show that images generated by our method are preferred by the users. The spread in the values computed with bootstrapping is $<$ 2.5\%.   
    }}
    \lbltbl{user_study}
    \vspace*{-5mm}
\end{table}

%% file: sections/5_discussion.tex
\vspace{-1mm}
\section{Conclusion and Limitations} \lblsec{discussion} \vspace{-2mm}
Our key idea is that different regions of an image are best inverted using different latent layers. We use this insight to train networks that predict the ``inversion difficulty" of different latent layers for any given input image. Image regions that are easy to reconstruct can be inverted using early latent layers, whereas difficult image regions should use the more capable feature space of the intermediate layers. We show inversion and editing results using our proposed multilayer inversion method on multiple challenging datasets. A limitation of this approach is that if a given input image is extremely difficult, our method will predict the use of the later latent layer that will correspond to being able to edit only limited things.

\vspace{-2mm}
\myparagraph{Acknowledgments.} 
We thank Eli Shechtman, Sheng-Yu Wang, Nupur Kumari, Kangle Deng, Muyang Li,  Bingliang Zhang, George Cazenavette, Ruihan Gao, and Chonghyuk (Andrew) Song  for useful discussions. We are
grateful for the support from Adobe, Naver Corporation, and Sony
Corporation.

%% file: sections/6_appendix.tex
\section{Appendix}

\subsection{Invertibility Prediction Network} 
\updates{First, we describe the architecture of the invertibility prediction network $S$, which is implemented as multi head network, and depicted in \reffig{arch_invertibility}. First is the shared base network $S_\text{base}$ that takes the target image $x$ as input and produces an intermediate output $S_\text{base}(x)$. The base network $S_\text{base}$ follows the design of ResNet DeepLabV3 ~\cite{chen2018encoder} with 8 output channels. Next, a separate head network $\text{head}_{\phi}$ is used for every candidate latent layer $\phi \in \{w^+, \Delta f_4, \Delta f_6, \Delta f_8, \Delta f_{10} \}$ and predicts the final invertibility for the corresponding latent layer $\phi$. Each head network is a 3 layer convolutional neural network with BatchNorm and LeakyReLU non linearity, except the last layer which uses ReLU. The base network and each of the head networks are trained jointly using the standard $l_2$ objective.
}

\input{figures/fig_apxx_inv_arch}

\subsection{Encoder Training}
In the \reffig{inv_train} we outlined how the learnt invertibility map can used for inverting StyleGAN2 models in an optimization framework. In this section we outline how the learnt invertibility can be used for training an encoder. 

\myparagraph{Encoder architecture details.} 
Next, we describe how our encoders for inversion are trained. 
We learn a different encoder for each of the latent spaces $\phi \in \{w^+, \Delta f_4, \Delta f_6, \Delta f_8, \Delta f_{10} \}$. Each encoder takes a 4 channel input - the input image concatenated with the corresponding binary mask. 
The encoder that predict the vector $w^+$ is initialized with a pretrained e4e encoder. The encoders corresponding to other latent spaces ($F_4, F_6, F_8, F_{10}$) borrow the ResNet-18 architecture with three changes: 1) The first convolution layer is modified to have 4 channels 2) We only retain the ResNet-18 layers till it matches the spatial resolution corresponding to the latent space its inverting. E.g. Encoder for $F_4$ will retain layers till $16 \times 16$ resolution.  3) Finally, we add a convolution layer to match the number of channels of the latent space.

\subsection{BigGAN-deep Inversion}
The \reffig{biggan_results} showed that the proposed spatially adaptive multilayer inversion can be applied to a class conditional model such as BigGAN-deep ~\cite{brock2019large}. Here we provide the details about the image formation model and the inversion objectives used. 
\myparagraph{Latent Spaces.} The BigGAN-deep generator takes a latent vector $z \in Z$ and a one-hot class label vector $c$ as the input. The latent vector $z$ is cloned and given as to multiple intermediate layers of the generator. We refer to this extended latent as $z^+ \in Z^+$. Next, the feature spaces used $F \in \{F_2, F_4\}$ follow the same definition mentioned in \refsec{approach}, and we similarly predict the change in the values $\Delta f$ instead of the values itself. Note that we do not optimize the class label in our experiments. Instead, we use a pretrained ResNet-50 ~\cite{he2016deep} classifier to predict the class label and keep it fixed throughout the inversion. 

\myparagraph{Inversion Training Objectives.} 
Next, we present the objective functions to optimize the latent code $\phi = \{ z^+, \Delta F_2, \Delta F_4\}$. We use the reconstruction loss $L_\text{rec}$ and F-space regularization $L_F$ defined in \refsec{approach}. For $z^+$, we use the following:
 \begin{equation}
\mathcal{L}_{Z}=\sum_n^N \big [ (z_n-\mu)^T \Sigma (z_n-\mu) + ||z_n^+ -z_0^+||^2 \big ], 
\end{equation}
where $z_n^+$ is the $n^\text{th}$ component of the $z^+$ vector, $\mu$ and $\Sigma$ are the empirical mean and covariance matrix of randomly sampled $z$ vectors respectively. The full objective is written as follows:
\begin{equation}
\arg \min_{\phi}  \mathcal{L}_{\text{rec}} + \lambda_{Z} \mathcal{L}_{Z} + \lambda_F \mathcal{L}_{F},
\end{equation}
\noindent where  $\lambda_{Z}$ and $\lambda_F$ control the weights for each term.

\subsection{Edit Directions}
Different style layers in the StyleGAN2 generator control different attributes in the generated image. Correspondingly, the latent space of an image segment determines whether a particular edit can be applied. \reftbl{edits_list} lists all edit directions used for our results and the corresponding latent spaces where they can be applied. $\textbf{W}^+$ is the most editable and image segments inverted with it can be edited with all edit directions. In contrast, only two of the edit directions can be applied for image segments inverted with $\text{F}_{10}$. 

\input{tables/ap1_edits}

\subsection{Ablation Studies}
\myparagraph{Regularizing the latent spaces.}
The equation 7 in the main paper regularizes the feature space to ensure our predicted feature values do not deviate too much from original feature space distribution. This is necessary to ensure that our edited direction is compatible with our predicted feature values. In \reffig{freg}, we show two inversion and editing results on the same input image. In the top row, the inversion is performed without the regularization $\mathcal{L}_F$ and the bottom row shows the results when the feature space is regularized. The images in the right column show the edited images when an edit direction corresponding to adding trees is applied with the \textit{same magnitude} to both the inverted latent codes. Both the cases achieve a good inversion, but the lack of feature space regularization results in an inversion that is not editable.
\input{figures/fig_ap1_freg}

\myparagraph{Refining the invertibility map.}
Next, we show the importance of the refinement step introduced in Section 3.1 of the main paper. This step ensures that a semantic class gets assigned to a single latent space. In \reffig{refine}, we show that edited image become inharmonious without such a refinement step. In particular, the top row in \reffig{refine} shows that different parts of the car get inverted using different latent spaces without the refinement step.
As a result, we get incoherent size change edits for the car as the edits will be applied to different layers for different car regions depending on which latent space it uses. 
\input{figures/fig_ap2_refining}

\myparagraph{Class segmentation as invertibility map.}
The well-performing GAN models are typically trained using an object specific dataset. Here, we consider an inversion approach that uses per-pixel class labels instead of our predicted invertibility map. For a GAN trained on the car images, we invert the car segment with $\text{W}^+$ and the rest of the image background is inverted using $\text{F}_6$ in \reffig{carseg}. 
\input{figures/fig_ap3_carseg}

\begin{table}[t!]
    \centering
    \resizebox{\linewidth}{!}{
    \begin{tabular}{l | c|c}
        \toprule 
        \textbf{Method} & \textbf{LPIPS-alex} & \textbf{LPIPS-vgg} \\
        \midrule
        
        e4e ~\cite{tov2021designing}            & \checkmark & - \\
        ReStyle (pSp) ~\cite{alaluf2021restyle} & \checkmark & - \\
        ReStyle (e4e) ~\cite{alaluf2021restyle} & \checkmark & - \\
        StyleGAN2 Inversion (W) ~\cite{karras2020analyzing} & - & \checkmark \\
        StyleGAN2 Inversion (W+) ~\cite{karras2020analyzing} & - & \checkmark \\
        Pivotal Tuning Inversion ~\cite{roich2021pivotal} & \checkmark & \checkmark \\
        SAMI - encoder (ours)      & - & \checkmark\\
        SAMI - optimization (ours) & - & \checkmark\\
        
        \bottomrule 
        
    \end{tabular}}
    \vspace{-1mm}
    \caption{\textbf{LPIPS backbone used by different inversion methods.}  }
    \lbltbl{lpips_types}
\end{table}

\begin{figure*}[t]
    \centering
    \includegraphics[width=0.95\linewidth]{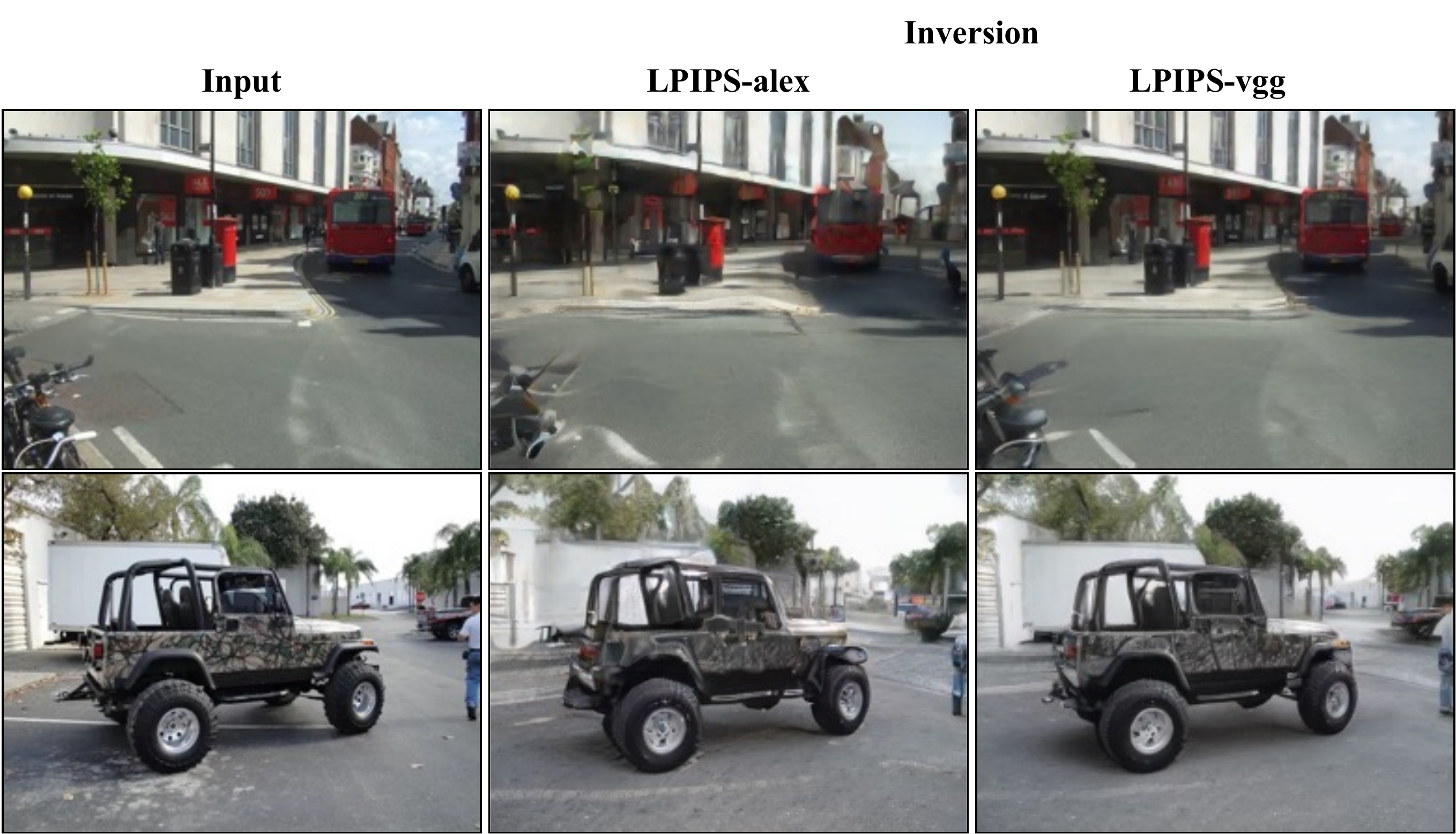}
    \caption{
    \textbf{Inversion with different perceptual metrics. } We invert input car images in two  settings using the objective function shown in Eq 8 in the main paper. Both the settings follow an identical inversion parameters except the architecture used for the LPIPS objective. The middle column shows the inverted image when LPIPS with AlexNet is used and the right column shows the inverted image when LPIPS with VGG is used. }
    \lblfig{lpips_type}
    \vspace{-5mm}
\end{figure*}

\subsection{Choosing the Perceptual Metric (LPIPS)}
\myparagraph{LPIPS-vgg vs. LPIPS-alex as perceptual loss.}
In \reffig{lpips_type} we show inversions of the same input image using both the variants of the LPIPS perceptual loss. LPIPS with VGG architecture results in an inversion that is sharper and has fewer artifacts. For instance, the inversion using LPIPS-alex is unable to reconstruct details such the the top of the red bus, the window pane (top row), the paint texture on the car, and background trees (bottom row). Inversion using LPIPS-vgg (right column) is able to reconstruct such details better. 

\myparagraph{Evaluation using different LPIPS architectures.}
Different inversion methods have used different LPIPS variants as their objective functions. \reftbl{lpips_types} enumerates the LPIPS versions used by different prior works. The use of the different LPIPS architectures by different prior inversion methods makes the evaluation across methods challenging.

\myparagraph{Societal Impacts.} Our approach can be useful for creative and media industries. Designers and artists can use our approach to perform meaningful edits on complex images which were possible to only limited extent using previous approaches. But there is always a danger of people using our approach to generate fake images to spread lie and propaganda on public forums. In order to counter that it would be important to use state-of-the-art fake image detection works~\cite{wang-cvpr2020}.

\input{figures/fig_apxx_cars_opt}

%% file: figures/fig_apxx_inv_arch.tex
\begin{figure*}[t]
    \centering
    \includegraphics[width=0.99\linewidth]{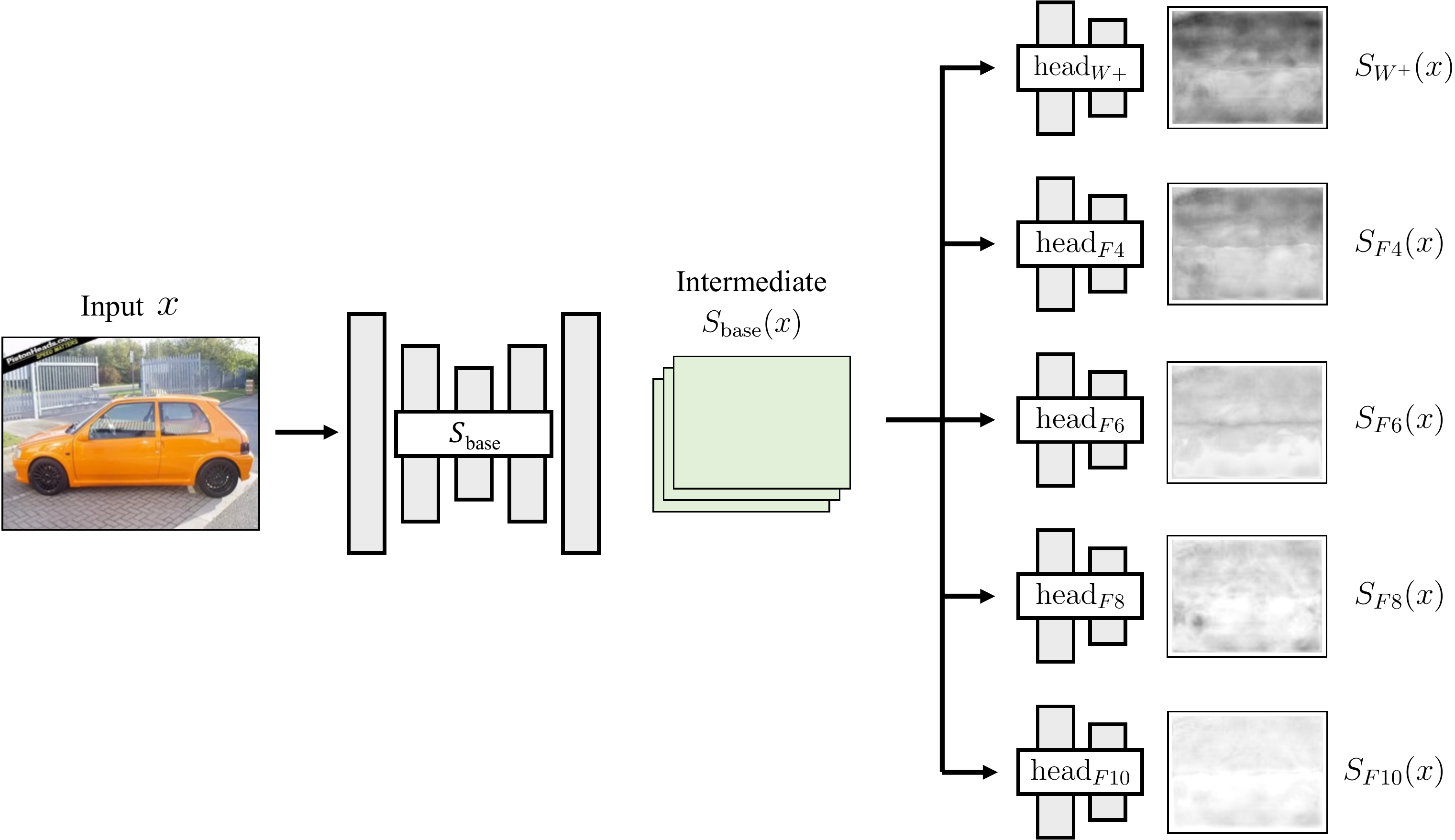}
    \caption{\updates{
    \textbf{Invertibility Prediction Architecture}. We show the architecture of the invertibility prediction network $S$ which comprises a shared backbone $S_{\text{base}}$, and individual head networks $\text{head}_{\phi_i}$ for each latent space $\phi_i$. The shared $S_{\text{base}}$ is implemented as a DeepLabv3 ~\cite{chen2018encoder} with a ResNet backbone and the head networks are 3 layer convolutional networks with BatchNorm and LeakyReLU nonlinearities.  
    }}
    \lblfig{arch_invertibility}
\end{figure*}

%% file: tables/ap1_edits.tex
\begin{table}[t!]
    \centering
    \resizebox{\linewidth}{!}{
    \begin{tabular}{c l | c c c c c}
        \toprule
        \textbf{Dataset} & \textbf{Direction Name} & \textbf{$\text{W}^+$} & \textbf{$\text{F}_4$} & \textbf{$\text{F}_6$} & \textbf{$\text{F}_8$} & \textbf{$\text{F}_{10}$} \\
        \midrule
        \multirow{4}{*}{LSUN - Cars} & car size & \checkmark & - & - & - & - \\
        & add trees & \checkmark & \checkmark & - & - & - \\
        & wheel type & \checkmark & \checkmark & \checkmark & - & - \\
        & car color (red) & \checkmark & \checkmark & \checkmark & \checkmark & \checkmark \\
        
        \cdashline{1-7}[4pt/2pt]
        \multirow{4}{*}{LSUN - Cats} & change pose & \checkmark & - & - & - & - \\
        & large cat eyes & \checkmark & \checkmark & \checkmark & - & - \\
        & cat fur color (black) & \checkmark & \checkmark & \checkmark & 
        \checkmark & - \\
        & cat with red nose & \checkmark & \checkmark & \checkmark &  \checkmark & - \\
        
        \cdashline{1-7}[4pt/2pt]
        \multirow{4}{*}{LSUN - Horses} & change pose & \checkmark & - & - & - & - \\
        & horse with a saddle & \checkmark & - & - & - & - \\
        & white horse & \checkmark & \checkmark & - & - & - \\
        & reduce trees & \checkmark & \checkmark & \checkmark & - & - \\
        
        \cdashline{1-7}[4pt/2pt]
        \multirow{4}{*}{FFHQ} & add glasses & \checkmark & - & - & - & - \\
        & laughing person & \checkmark & \checkmark & - & - & - \\
        & thick eyebrows & \checkmark & \checkmark & \checkmark & \checkmark & - \\
        & increase age & \checkmark & \checkmark & \checkmark & \checkmark & \checkmark \\
        \bottomrule 
    \end{tabular}}
    \caption{\textbf{Editing ability of different latent layers.} We enumerate all the edit directions for StyleGAN2 trained on different datasets, and show the latent spaces that can be used for editing them. For each editing direction, every latent space capable of performing this edit is labeled by a \checkmark. For instance, the car color can be controlled by every latent space and correspondingly all the row entries are marked by a \checkmark. In contrast, the car size can only be controlled by $\text{W}^+$.} 
    \lbltbl{edits_list}
    \vspace*{-.15in}
\end{table}

%% file: figures/fig_ap1_freg.tex
\begin{figure*}[t]
    \centering
    \includegraphics[width=0.95\linewidth]{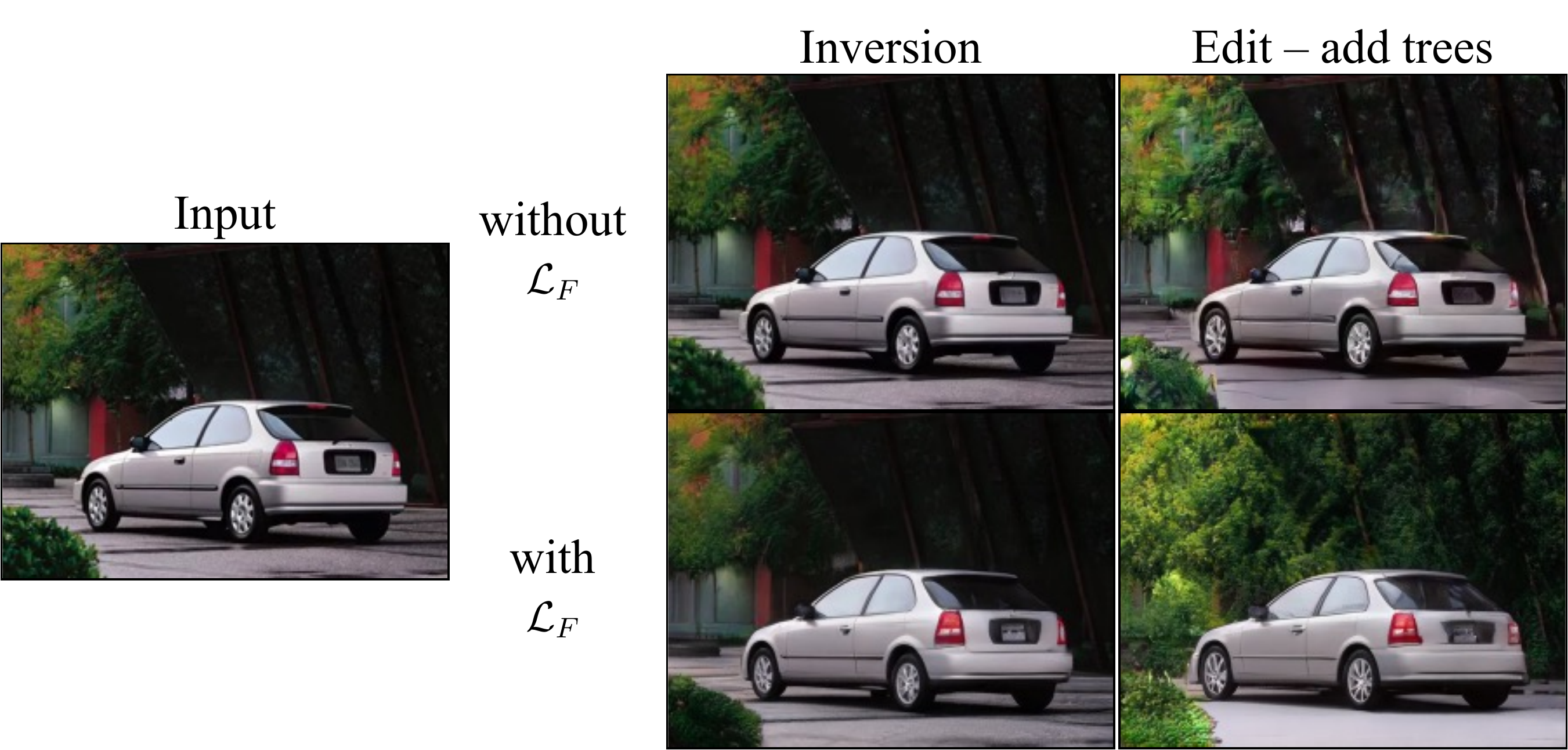}
    \caption{
    \textbf{Regularization in the features space F. } In top row, we show results without feature space regularization. We can see that edit to add tree does not work well without regularization as our predicted feature space may not be close original feature space distribution and edit direction would not be compatible anymore. }
    \lblfig{freg}
    \vspace{-5mm}
\end{figure*}

%% file: figures/fig_ap2_refining.tex
\begin{figure*}[t]
    \centering
    \includegraphics[width=0.95\linewidth]{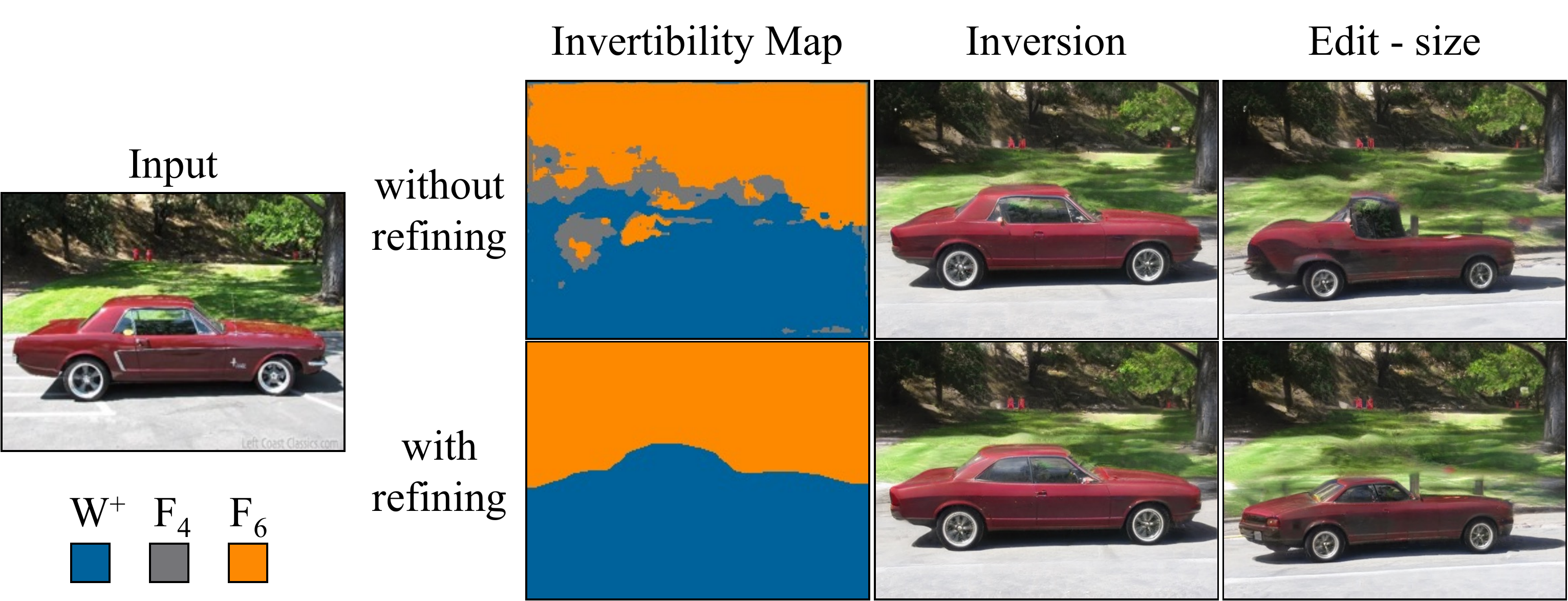}
    \caption{
    \textbf{Necessity for refining the invertibility map.} In top row, without refining the car segment get assigned to multiple feature space which results in inconsistent edit with artifacts. Where as with refining in bottom row, the entire car region get assigned to $\text{W}^{+}$ space which gives us consistent edit of changing the car size.  }
    \lblfig{refine}
    \vspace{-5mm}
\end{figure*}

%% file: figures/fig_ap3_carseg.tex
\begin{figure*}[t]
    \vspace{5mm}
    \centering
    \includegraphics[width=0.99\linewidth]{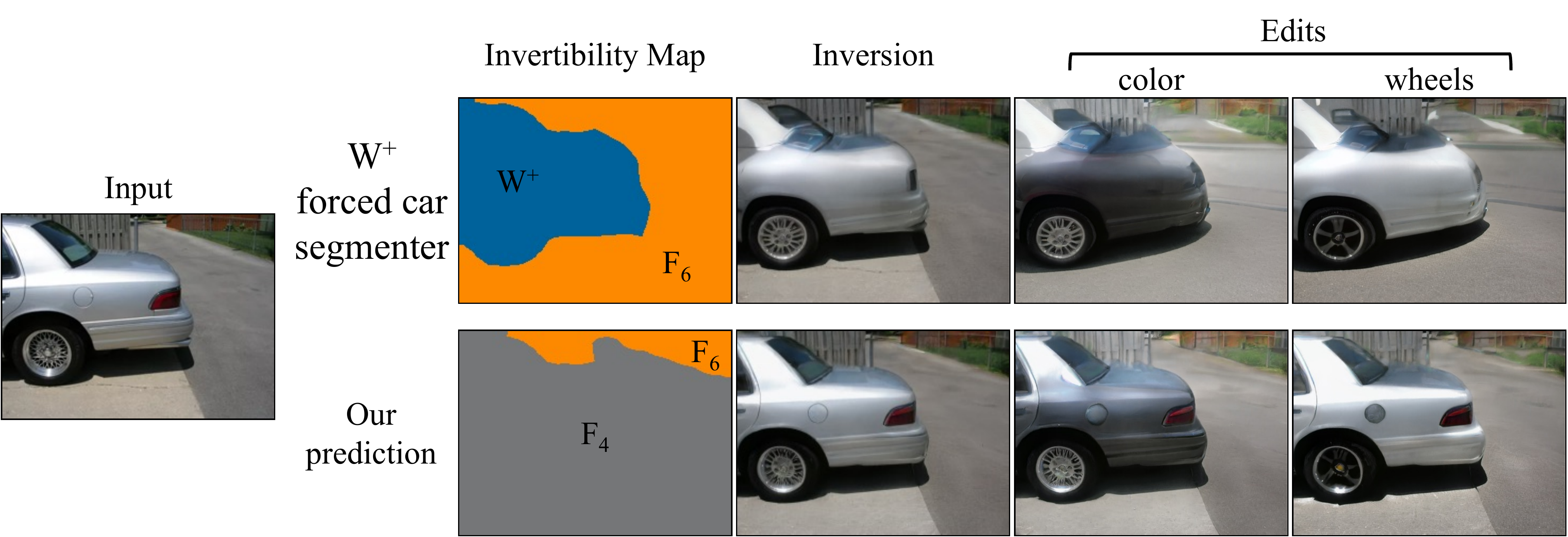}
    \vspace{-3mm}
    \caption{
    \textbf{Using a car segmenter.} We first invert a given target image with the assumption that the car regions of the image should be invertible with the native $\text{W}^{+}$. This assumption leads to a poor inversion that is not able to reconstruct the target car image. Whereas our method correctly predicts a good latent space for the different regions and consequently generated better inversions and edits which retain the identity of the original car better.
    }
    \lblfig{carseg}
    \vspace{-.15in}
\end{figure*}

%% file: figures/fig_apxx_cars_opt.tex
\begin{figure*}[t]
    \centering
    \includegraphics[width=0.99\linewidth]{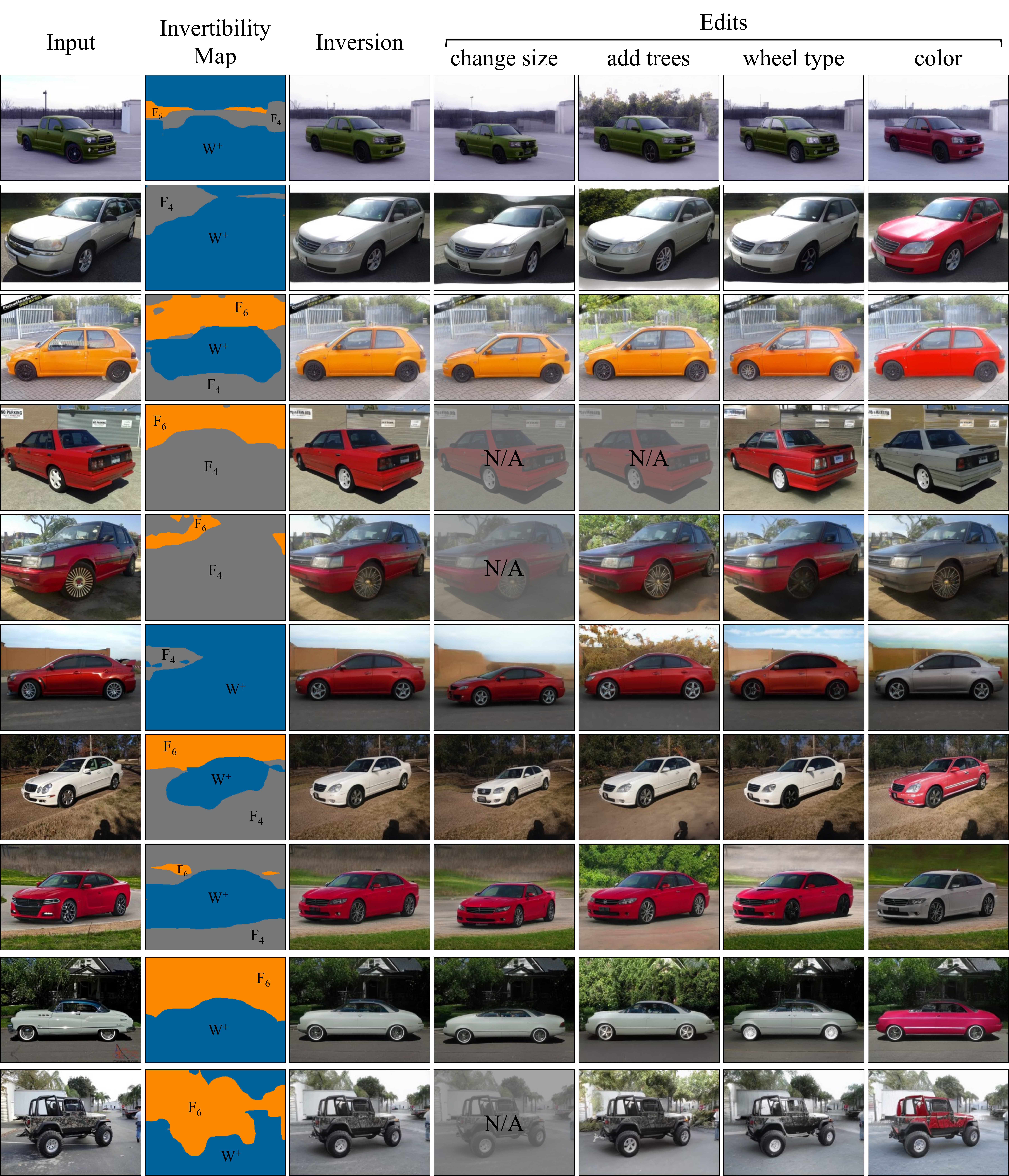}
    \caption{
    Additional inversion and editing results with the proposed SAM method on cars.
    }
    \lblfig{grid_cars_opt}
\end{figure*}

\begin{figure*}[t]
    \centering
    \includegraphics[width=0.99\linewidth]{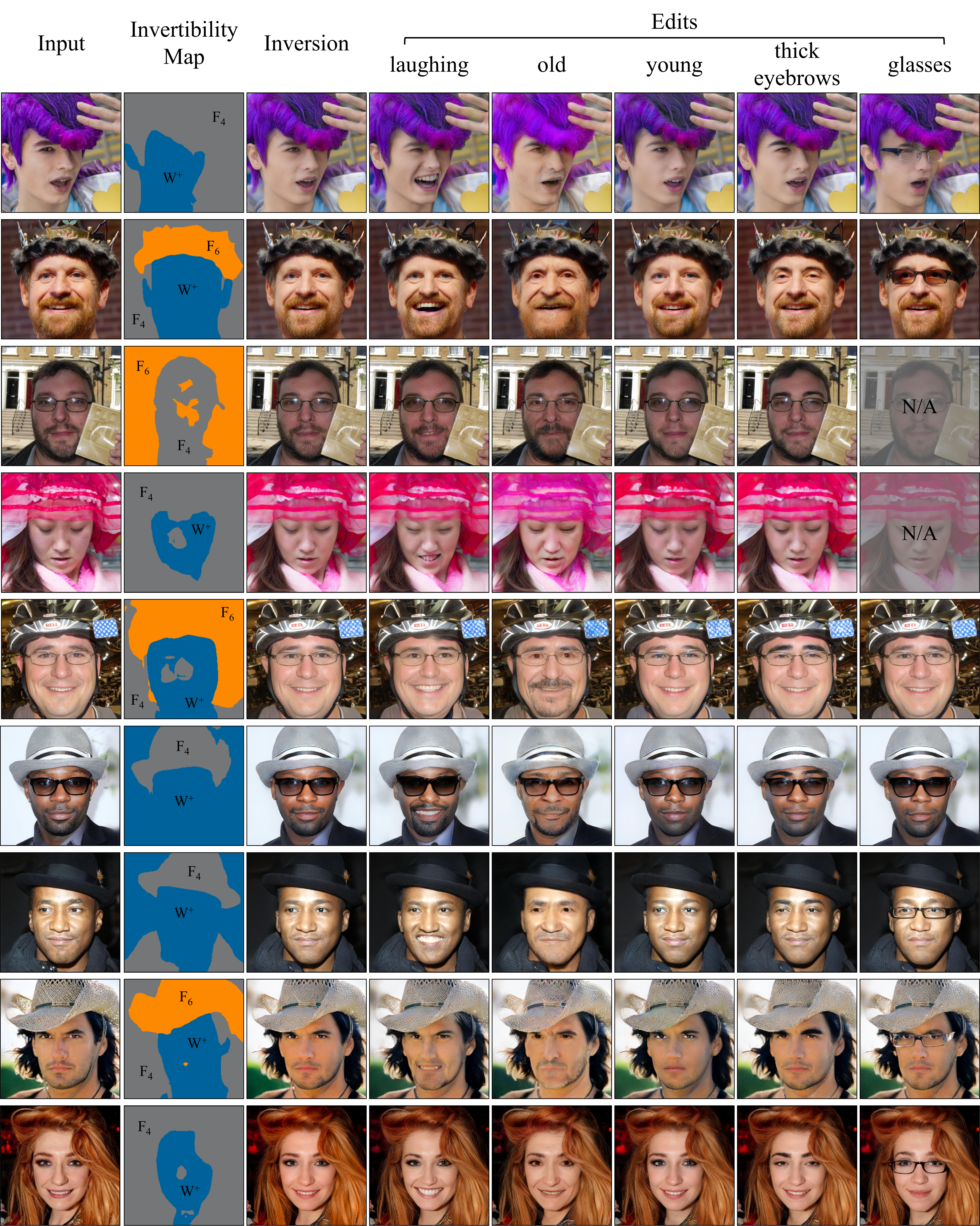}
    \caption{
    Additional inversion and editing results with the proposed SAM method on challenging faces.
    }
    \lblfig{grid_faces_opt}
\end{figure*}

%% file: main_arxiv.bbl
\begin{thebibliography}{10}\itemsep=-1pt

\bibitem{abdal2019image2stylegan}
Rameen Abdal, Yipeng Qin, and Peter Wonka.
\newblock Image2stylegan: How to embed images into the stylegan latent space?
\newblock In {\em IEEE International Conference on Computer Vision (ICCV)},
  2019.

\bibitem{abdal2020image2stylegan++}
Rameen Abdal, Yipeng Qin, and Peter Wonka.
\newblock Image2stylegan++: How to edit the embedded images?
\newblock In {\em IEEE Conference on Computer Vision and Pattern Recognition
  (CVPR)}, 2020.

\bibitem{abdal2021styleflow}
Rameen Abdal, Peihao Zhu, Niloy~J Mitra, and Peter Wonka.
\newblock Styleflow: Attribute-conditioned exploration of stylegan-generated
  images using conditional continuous normalizing flows.
\newblock {\em ACM Transactions on Graphics (TOG)}, 40(3):1--21, 2021.

\bibitem{alaluf2021matter}
Yuval Alaluf, Or Patashnik, and Daniel Cohen-Or.
\newblock Only a matter of style: Age transformation using a style-based
  regression model.
\newblock {\em ACM Trans. Graph.}, 40(4), 2021.

\bibitem{alaluf2021restyle}
Yuval Alaluf, Or Patashnik, and Daniel Cohen-Or.
\newblock Restyle: A residual-based stylegan encoder via iterative refinement.
\newblock In {\em Proceedings of the IEEE/CVF International Conference on
  Computer Vision (ICCV)}, October 2021.

\bibitem{anirudh2020mimicgan}
Rushil Anirudh, Jayaraman~J Thiagarajan, Bhavya Kailkhura, and Peer-Timo
  Bremer.
\newblock Mimicgan: Robust projection onto image manifolds with corruption
  mimicking.
\newblock {\em International Journal of Computer Vision}, pages 1--19, 2020.

\bibitem{asim2018blind}
Muhammad Asim, Fahad Shamshad, and Ali Ahmed.
\newblock Blind image deconvolution using deep generative priors.
\newblock {\em arXiv preprint arXiv:1802.04073}, 2018.

\bibitem{bau2019semantic}
David Bau, Hendrik Strobelt, William Peebles, Jonas Wulff, Bolei Zhou, Jun-Yan
  Zhu, and Antonio Torralba.
\newblock Semantic photo manipulation with a generative image prior.
\newblock {\em ACM SIGGRAPH}, 38(4):1--11, 2019.

\bibitem{bau2019gandissect}
David Bau, Jun-Yan Zhu, Hendrik Strobelt, Bolei Zhou, Joshua~B. Tenenbaum,
  William~T. Freeman, and Antonio Torralba.
\newblock Gan dissection: Visualizing and understanding generative adversarial
  networks.
\newblock In {\em International Conference on Learning Representations (ICLR)},
  2019.

\bibitem{bau2019seeing}
David Bau, Jun-Yan Zhu, Jonas Wulff, William Peebles, Hendrik Strobelt, Bolei
  Zhou, and Antonio Torralba.
\newblock Seeing what a gan cannot generate.
\newblock In {\em IEEE International Conference on Computer Vision (ICCV)},
  2019.

\bibitem{binkowski2018demystifying}
Miko{\l}aj Bi{\'n}kowski, Danika~J Sutherland, Michael Arbel, and Arthur
  Gretton.
\newblock Demystifying mmd gans.
\newblock In {\em ICLR}, 2018.

\bibitem{brock2019large}
Andrew Brock, Jeff Donahue, and Karen Simonyan.
\newblock Large scale gan training for high fidelity natural image synthesis.
\newblock In {\em International Conference on Learning Representations (ICLR)},
  2019.

\bibitem{brock2016neural}
Andrew Brock, Theodore Lim, James~M Ritchie, and Nick Weston.
\newblock Neural photo editing with introspective adversarial networks.
\newblock In {\em International Conference on Learning Representations (ICLR)},
  2017.

\bibitem{brock2017neural}
Andrew Brock, Theodore Lim, James~M Ritchie, and Nick Weston.
\newblock Neural photo editing with introspective adversarial networks.
\newblock In {\em International Conference on Learning Representations (ICLR)},
  2017.

\bibitem{chai2021using}
Lucy Chai, Jonas Wulff, and Phillip Isola.
\newblock Using latent space regression to analyze and leverage
  compositionality in gans.
\newblock In {\em International Conference on Learning Representations (ICLR)},
  2021.

\bibitem{chai2021ensembling}
Lucy Chai, Jun-Yan Zhu, Eli Shechtman, Phillip Isola, and Richard Zhang.
\newblock Ensembling with deep generative views.
\newblock In {\em Proceedings of the IEEE/CVF Conference on Computer Vision and
  Pattern Recognition}, pages 14997--15007, 2021.

\bibitem{chen2018encoder}
Liang-Chieh Chen, Yukun Zhu, George Papandreou, Florian Schroff, and Hartwig
  Adam.
\newblock Encoder-decoder with atrous separable convolution for semantic image
  segmentation.
\newblock In {\em Proceedings of the European conference on computer vision
  (ECCV)}, pages 801--818, 2018.

\bibitem{collins2020editing}
Edo Collins, Raja Bala, Bob Price, and Sabine Susstrunk.
\newblock Editing in style: Uncovering the local semantics of gans.
\newblock In {\em IEEE Conference on Computer Vision and Pattern Recognition
  (CVPR)}, 2020.

\bibitem{futschik2021real}
David Futschik, Michal Luk{\'a}{\v{c}}, Eli Shechtman, and Daniel S{\`y}kora.
\newblock Real image inversion via segments.
\newblock {\em arXiv preprint arXiv:2110.06269}, 2021.

\bibitem{goodfellow2014generative}
Ian Goodfellow, Jean Pouget-Abadie, Mehdi Mirza, Bing Xu, David Warde-Farley,
  Sherjil Ozair, Aaron Courville, and Yoshua Bengio.
\newblock Generative adversarial nets.
\newblock In {\em Advances in Neural Information Processing Systems}, 2014.

\bibitem{gu2020image}
Jinjin Gu, Yujun Shen, and Bolei Zhou.
\newblock Image processing using multi-code gan prior.
\newblock In {\em IEEE Conference on Computer Vision and Pattern Recognition
  (CVPR)}, 2020.

\bibitem{harkonen2020ganspace}
Erik H{\"a}rk{\"o}nen, Aaron Hertzmann, Jaakko Lehtinen, and Sylvain Paris.
\newblock Ganspace: Discovering interpretable gan controls.
\newblock In {\em Advances in Neural Information Processing Systems}, 2020.

\bibitem{he2016deep}
Kaiming He, Xiangyu Zhang, Shaoqing Ren, and Jian Sun.
\newblock Deep residual learning for image recognition.
\newblock In {\em IEEE Conference on Computer Vision and Pattern Recognition
  (CVPR)}, 2016.

\bibitem{heusel2017gans}
Martin Heusel, Hubert Ramsauer, Thomas Unterthiner, Bernhard Nessler, and Sepp
  Hochreiter.
\newblock {GANs} trained by a two time-scale update rule converge to a local
  {Nash} equilibrium.
\newblock In {\em Advances in Neural Information Processing Systems}, 2017.

\bibitem{huang2017adain}
Xun Huang and Serge Belongie.
\newblock Arbitrary style transfer in real-time with adaptive instance
  normalization.
\newblock In {\em IEEE International Conference on Computer Vision (ICCV)},
  2017.

\bibitem{huh2020transforming}
Minyoung Huh, Richard Zhang, Jun-Yan Zhu, Sylvain Paris, and Aaron Hertzmann.
\newblock Transforming and projecting images into class-conditional generative
  networks.
\newblock In {\em European Conference on Computer Vision (ECCV)}, 2020.

\bibitem{gansteerability}
Ali Jahanian, Lucy Chai, and Phillip Isola.
\newblock On the "steerability" of generative adversarial networks.
\newblock In {\em International Conference on Learning Representations}, 2020.

\bibitem{kafri2021stylefusion}
Omer Kafri, Or Patashnik, Yuval Alaluf, and Daniel Cohen-Or.
\newblock Stylefusion: A generative model for disentangling spatial segments.
\newblock {\em arXiv preprint arXiv:2107.07437}, 2021.

\bibitem{kang2021gan}
Kyoungkook Kang, Seongtae Kim, and Sunghyun Cho.
\newblock Gan inversion for out-of-range images with geometric transformations.
\newblock In {\em IEEE International Conference on Computer Vision (ICCV)},
  2021.

\bibitem{karnewar2020msg}
Animesh Karnewar and Oliver Wang.
\newblock Msg-gan: Multi-scale gradients for generative adversarial networks.
\newblock In {\em IEEE Conference on Computer Vision and Pattern Recognition
  (CVPR)}, 2020.

\bibitem{karras2018progressive}
Tero Karras, Timo Aila, Samuli Laine, and Jaakko Lehtinen.
\newblock Progressive growing of gans for improved quality, stability, and
  variation.
\newblock In {\em International Conference on Learning Representations (ICLR)},
  2018.

\bibitem{karras2020training}
Tero Karras, Miika Aittala, Janne Hellsten, Samuli Laine, Jaakko Lehtinen, and
  Timo Aila.
\newblock Training generative adversarial networks with limited data.
\newblock {\em NIPS}, 33, 2020.

\bibitem{karras2021alias}
Tero Karras, Miika Aittala, Samuli Laine, Erik H{\"a}rk{\"o}nen, Janne
  Hellsten, Jaakko Lehtinen, and Timo Aila.
\newblock Alias-free generative adversarial networks.
\newblock {\em arXiv preprint arXiv:2106.12423}, 2021.

\bibitem{karras2019style}
Tero Karras, Samuli Laine, and Timo Aila.
\newblock A style-based generator architecture for generative adversarial
  networks.
\newblock In {\em IEEE Conference on Computer Vision and Pattern Recognition
  (CVPR)}, 2019.

\bibitem{karras2020analyzing}
Tero Karras, Samuli Laine, Miika Aittala, Janne Hellsten, Jaakko Lehtinen, and
  Timo Aila.
\newblock Analyzing and improving the image quality of stylegan.
\newblock In {\em IEEE Conference on Computer Vision and Pattern Recognition
  (CVPR)}, 2020.

\bibitem{kim2021exploiting}
Hyunsu Kim, Yunjey Choi, Junho Kim, Sungjoo Yoo, and Youngjung Uh.
\newblock Exploiting spatial dimensions of latent in gan for real-time image
  editing.
\newblock In {\em IEEE Conference on Computer Vision and Pattern Recognition
  (CVPR)}, 2021.

\bibitem{larsen2016autoencoding}
Anders Boesen~Lindbo Larsen, S{\o}ren~Kaae S{\o}nderby, Hugo Larochelle, and
  Ole Winther.
\newblock Autoencoding beyond pixels using a learned similarity metric.
\newblock In {\em International Conference on Machine Learning (ICML)}, 2016.

\bibitem{lipton2017precise}
Zachary~C Lipton and Subarna Tripathi.
\newblock Precise recovery of latent vectors from generative adversarial
  networks.
\newblock {\em arXiv preprint arXiv:1702.04782}, 2017.

\bibitem{luo2020time}
Xuan Luo, Xuaner Zhang, Paul Yoo, Ricardo Martin-Brualla, Jason Lawrence, and
  Steven~M Seitz.
\newblock Time-travel rephotography.
\newblock {\em arXiv preprint arXiv:2012.12261}, 2020.

\bibitem{pan2021exploiting}
Xingang Pan, Xiaohang Zhan, Bo Dai, Dahua Lin, Chen~Change Loy, and Ping Luo.
\newblock Exploiting deep generative prior for versatile image restoration and
  manipulation.
\newblock {\em IEEE Transactions on Pattern Analysis and Machine Intelligence
  (TPAMI)}, 2021.

\bibitem{park2020swapping}
Taesung Park, Jun-Yan Zhu, Oliver Wang, Jingwan Lu, Eli Shechtman, Alexei~A.
  Efros, and Richard Zhang.
\newblock Swapping autoencoder for deep image manipulation.
\newblock In {\em Advances in Neural Information Processing Systems}, 2020.

\bibitem{patashnik2021styleclip}
Or Patashnik, Zongze Wu, Eli Shechtman, Daniel Cohen-Or, and Dani Lischinski.
\newblock Styleclip: Text-driven manipulation of stylegan imagery.
\newblock In {\em Proceedings of the IEEE/CVF International Conference on
  Computer Vision (ICCV)}, 2021.

\bibitem{peebles2020hessian}
William Peebles, John Peebles, Jun-Yan Zhu, Alexei Efros, and Antonio Torralba.
\newblock The hessian penalty: A weak prior for unsupervised disentanglement.
\newblock In {\em European Conference on Computer Vision (ECCV)}. Springer,
  2020.

\bibitem{perarnau2016invertible}
Guim Perarnau, Joost van~de Weijer, Bogdan Raducanu, and Jose~M {\'A}lvarez.
\newblock Invertible conditional gans for image editing.
\newblock In {\em NIPS Workshop on Adversarial Training}, 2016.

\bibitem{richardson2020encoding}
Elad Richardson, Yuval Alaluf, Or Patashnik, Yotam Nitzan, Yaniv Azar, Stav
  Shapiro, and Daniel Cohen-Or.
\newblock Encoding in style: a stylegan encoder for image-to-image translation.
\newblock {\em arXiv preprint arXiv:2008.00951}, 2020.

\bibitem{roich2021pivotal}
Daniel Roich, Ron Mokady, Amit~H Bermano, and Daniel Cohen-Or.
\newblock Pivotal tuning for latent-based editing of real images.
\newblock {\em arXiv preprint arXiv:2106.05744}, 2021.

\bibitem{russakovsky2015imagenet}
Olga Russakovsky, Jia Deng, Hao Su, Jonathan Krause, Sanjeev Satheesh, Sean Ma,
  Zhiheng Huang, Andrej Karpathy, Aditya Khosla, Michael Bernstein, et~al.
\newblock Imagenet large scale visual recognition challenge.
\newblock {\em International Journal of Computer Vision (IJCV)},
  115(3):211--252, 2015.

\bibitem{salimans2016improved}
Tim Salimans, Ian Goodfellow, Wojciech Zaremba, Vicki Cheung, Alec Radford, and
  Xi Chen.
\newblock Improved techniques for training gans.
\newblock In {\em Advances in Neural Information Processing Systems}, 2016.

\bibitem{shen2020interfacegan}
Yujun Shen, Ceyuan Yang, Xiaoou Tang, and Bolei Zhou.
\newblock Interfacegan: Interpreting the disentangled face representation
  learned by gans.
\newblock {\em IEEE Transactions on Pattern Analysis and Machine Intelligence
  (TPAMI)}, 2020.

\bibitem{shen2021closed}
Yujun Shen and Bolei Zhou.
\newblock Closed-form factorization of latent semantics in gans.
\newblock In {\em IEEE Conference on Computer Vision and Pattern Recognition
  (CVPR)}, 2021.

\bibitem{suzuki2018spatially}
Ryohei Suzuki, Masanori Koyama, Takeru Miyato, Taizan Yonetsuji, and Huachun
  Zhu.
\newblock Spatially controllable image synthesis with internal representation
  collaging.
\newblock {\em arXiv preprint arXiv:1811.10153}, 2018.

\bibitem{tewari2020stylerig}
Ayush Tewari, Mohamed Elgharib, Gaurav Bharaj, Florian Bernard, Hans-Peter
  Seidel, Patrick P{\'e}rez, Michael Zollhofer, and Christian Theobalt.
\newblock Stylerig: Rigging stylegan for 3d control over portrait images.
\newblock In {\em IEEE Conference on Computer Vision and Pattern Recognition
  (CVPR)}, 2020.

\bibitem{tov2021designing}
Omer Tov, Yuval Alaluf, Yotam Nitzan, Or Patashnik, and Daniel Cohen-Or.
\newblock Designing an encoder for stylegan image manipulation.
\newblock In {\em ACM SIGGRAPH}, 2021.

\bibitem{voynov2020unsupervised}
Andrey Voynov and Artem Babenko.
\newblock Unsupervised discovery of interpretable directions in the gan latent
  space.
\newblock In {\em International Conference on Machine Learning (ICML)}, 2020.

\bibitem{wang-cvpr2020}
Sheng-Yu Wang, Oliver Wang, Richard Zhang, Andrew Owens, and Alexei~A Efros.
\newblock Cnn-generated images are surprisingly easy to spot...for now.
\newblock In {\em CVPR}, 2020.

\bibitem{wei2021simpleinversion}
Tianyi Wei, Dongdong Chen, Wenbo Zhou, Jing Liao, Weiming Zhang, Lu Yuan, Gang
  Hua, and Nenghai Yu.
\newblock A simple baseline for stylegan inversion.
\newblock {\em arXiv preprint arXiv:2104.07661}, 2021.

\bibitem{wu2021towards}
Yanze Wu, Xintao Wang, Yu Li, Honglun Zhang, Xun Zhao, and Ying Shan.
\newblock Towards vivid and diverse image colorization with generative color
  prior.
\newblock In {\em IEEE International Conference on Computer Vision (ICCV)},
  2021.

\bibitem{wu2020stylespace}
Zongze Wu, Dani Lischinski, and Eli Shechtman.
\newblock Stylespace analysis: Disentangled controls for stylegan image
  generation.
\newblock In {\em CVPR}, 2021.

\bibitem{wulff2020improving}
Jonas Wulff and Antonio Torralba.
\newblock Improving inversion and generation diversity in stylegan using a
  gaussianized latent space.
\newblock {\em arXiv preprint arXiv:2009.06529}, 2020.

\bibitem{xia2021gan}
Weihao Xia, Yulun Zhang, Yujiu Yang, Jing-Hao Xue, Bolei Zhou, and Ming-Hsuan
  Yang.
\newblock Gan inversion: A survey.
\newblock {\em arXiv preprint arXiv:2101.05278}, 2021.

\bibitem{xu2021continuity}
Yangyang Xu, Yong Du, Wenpeng Xiao, Xuemiao Xu, and Shengfeng He.
\newblock From continuity to editability: Inverting gans with consecutive
  images.
\newblock In {\em Proceedings of the IEEE/CVF International Conference on
  Computer Vision}, pages 13910--13918, 2021.

\bibitem{yeh2017semantic}
Raymond~A Yeh, Chen Chen, Teck Yian~Lim, Alexander~G Schwing, Mark
  Hasegawa-Johnson, and Minh~N Do.
\newblock Semantic image inpainting with deep generative models.
\newblock In {\em IEEE Conference on Computer Vision and Pattern Recognition
  (CVPR)}, 2017.

\bibitem{zhang2018unreasonable}
Richard Zhang, Phillip Isola, Alexei~A Efros, Eli Shechtman, and Oliver Wang.
\newblock The unreasonable effectiveness of deep features as a perceptual
  metric.
\newblock In {\em IEEE Conference on Computer Vision and Pattern Recognition
  (CVPR)}, 2018.

\bibitem{zhao2020diffaugment}
Shengyu Zhao, Zhijian Liu, Ji Lin, Jun-Yan Zhu, and Song Han.
\newblock Differentiable augmentation for data-efficient gan training.
\newblock In {\em Advances in Neural Information Processing Systems}, 2020.

\bibitem{zhu2020indomain}
Jiapeng Zhu, Yujun Shen, Deli Zhao, and Bolei Zhou.
\newblock In-domain gan inversion for real image editing.
\newblock In {\em ECCV}, 2020.

\bibitem{zhu2016generative}
Jun-Yan Zhu, Philipp Kr{\"a}henb{\"u}hl, Eli Shechtman, and Alexei~A Efros.
\newblock Generative visual manipulation on the natural image manifold.
\newblock In {\em European Conference on Computer Vision (ECCV)}, 2016.

\bibitem{zhu2021barbershop}
Peihao Zhu, Rameen Abdal, John Femiani, and Peter Wonka.
\newblock Barbershop: Gan-based image compositing using segmentation masks.
\newblock {\em arXiv preprint arXiv:2106.01505}, 2021.

\bibitem{zhu2020improved}
Peihao Zhu, Rameen Abdal, Yipeng Qin, John Femiani, and Peter Wonka.
\newblock Improved stylegan embedding: Where are the good latents?
\newblock {\em arXiv preprint arXiv:2012.09036}, 2020.

\end{thebibliography}
